\documentclass[manuscript,screen]{acmart}

\usepackage{multirow}
\usepackage{booktabs}
\usepackage{adjustbox}
\usepackage{listings}
\usepackage{graphicx}
\usepackage{subcaption}
\usepackage{tabularx}
\usepackage{soul}
\usepackage{hyperref}

\usepackage{graphicx}
\usepackage{amsmath}
\usepackage{amsthm}
\usepackage{booktabs}
\usepackage{algorithm}
\usepackage{algorithmic}

\usepackage{amssymb}
\usepackage{multirow}
\usepackage{boldline}
\usepackage{hhline}
\usepackage{xcolor}
\usepackage{pifont}
\usepackage{makecell}
\usepackage{bbm}

\usepackage{utfsym}
\usepackage[normalem]{ulem}
\useunder{\uline}{\ul}{}
\usepackage[inline]{enumitem}
\usepackage{enumitem}
\usepackage{IEEEtrantools}
\usepackage{stmaryrd}
\usepackage{array}
\definecolor{hidden-red}{RGB}{205, 44, 36}
\definecolor{hidden-blue}{RGB}{194,232,247}
\definecolor{hidden-orange}{RGB}{243,202,120}
\definecolor{hidden-green}{RGB}{34,139,34}
\definecolor{hidden-pink}{RGB}{255,245,247}
\definecolor{hidden-black}{RGB}{20,68,106}

\newcommand{\xmark}{\ding{55}}

\newcommand{\highlight}[1]{\textcolor{black}{#1}}

\usepackage[edges]{forest}

\AtBeginDocument{%
  }

\setcopyright{acmlicensed}
\copyrightyear{2018}
\acmYear{2018}
\acmDOI{XXXXXXX.XXXXXXX}

\acmConference[Conference acronym 'XX]{Make sure to enter the correct
  conference title from your rights confirmation emai}{June 03--05,
  2018}{Woodstock, NY}
\acmISBN{978-1-4503-XXXX-X/18/06}

\begin{document}

\title{Recent Advances of Foundation Language Models-based Continual Learning: A Survey}


\author{Yutao Yang}

\author{Jie Zhou}
\authornote{Corresponding authors.}
\email{jzhou@cs.ecnu.edu.cn}

\author{Xuanwen Ding}

\author{Tianyu Huai}

\author{Shunyu Liu}

\author{Qin Chen}

\author{Yuan Xie}

\author{Liang He}
\affiliation{%
  \institution{School of Computer Science and Technology, East China Normal University}
  \city{Shanghai}
  \country{China}
}

\renewcommand{\shortauthors}{Yutao Yang et al.}

\begin{abstract}
Recently, foundation language models (LMs) have marked significant achievements in the domains of natural language processing (NLP) and computer vision (CV). Unlike traditional neural network models, foundation LMs obtain a great ability for transfer learning by acquiring rich commonsense knowledge through pre-training on extensive unsupervised datasets with a vast number of parameters. Despite these capabilities, LMs still struggle with catastrophic forgetting, hindering their ability to learn continuously like humans. To address this, continual learning (CL) methodologies have been introduced, allowing LMs to adapt to new tasks while retaining learned knowledge. However, a systematic taxonomy of existing approaches and a comparison of their performance are still lacking. In this paper, we delve into a comprehensive review, summarization, and classification of the existing literature on CL-based approaches applied to foundation language models, such as pre-trained language models (PLMs), large language models (LLMs) and vision-language models (VLMs). We divide these studies into offline and online CL, which consist of traditional methods, parameter-efficient-based methods, instruction tuning-based methods and continual pre-training methods. 
Additionally, we outline the typical datasets and metrics employed in CL research and provide a detailed analysis of the challenges and future work for LMs-based continual learning.
\end{abstract}




\maketitle

\section{Introduction}
\label{sect: Introduciton}

Recent advancements in foundation language models (LMs) have set new benchmarks in both natural language processing (NLP) \cite{zhao2023survey,min2023recent,zhou2023chatgpt} and computer vision (CV) \cite{wang2023large, tiwari2022gcr, ramasesh2022effect}. Foundation LMs encompass three primary categories: Pre-trained Language Models (PLMs) \cite{min2023recent}, Large Language Models (LLMs) \cite{zhao2023survey}, and Vision-Language Models (VLMs) \cite{ijcai2022p762}. 
\highlight{These models are pre-trained on large, unlabeled datasets to capture rich semantic information, which is then fine-tuned for specific tasks or domains. This strategy not only enhances performance across various applications but also significantly improves the flexibility and adaptability of these models.} 

\highlight{However, despite their strengths, foundation LMs face challenges in dynamic environments where tasks evolve over time. A key issue is ``catastrophic forgetting" \cite{kirkpatrick2017overcoming}, where models lose previously learned knowledge when adapting to new information. Unlike human learning, which is inherently continuous and adaptive \cite{de2021continual}, foundation LMs generally require retraining to incorporate new data. Effective learning in such environments demands not only the ability to accelerate learning on new tasks (forward transfer) but also to improve performance on previous tasks by integrating newly acquired knowledge (backward transfer). While multi-task learning (MTL) and transfer learning (TL) offer potential solutions, MTL requires all task data to be available upfront, and TL focuses on limited tasks, making both approaches impractical for dynamic, real-world applications.}



\begin{figure}
\vspace{-2mm}
    \centering
    \includegraphics[width=0.7\linewidth]{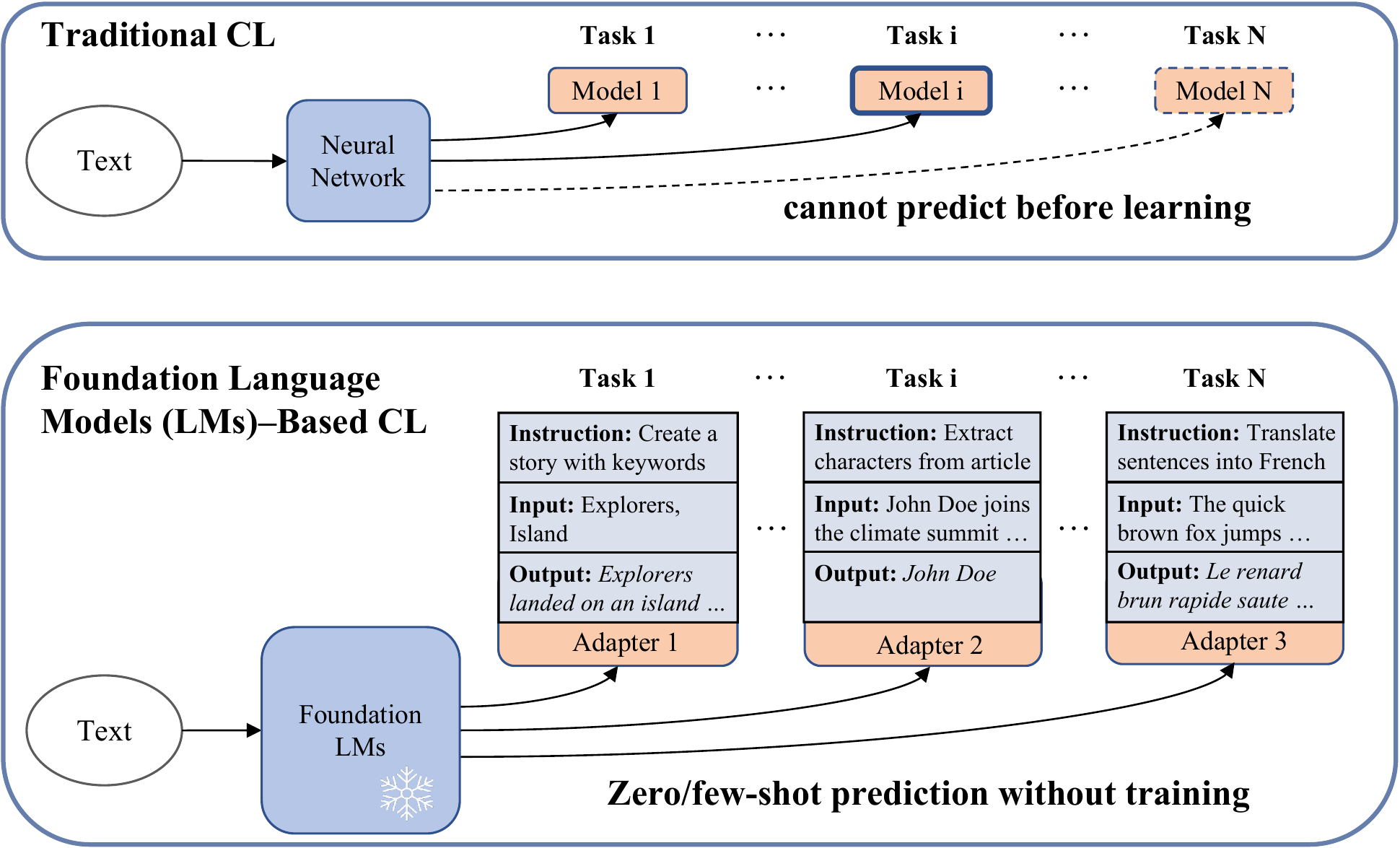}
    \vspace{-1mm}
    \caption{Comparison between traditional CL and Foundation language models (LMs)-Based CL.}
    \label{fig:IntroCLDifference}
    \vspace{-3mm}
\end{figure}

Continual learning (CL) \cite{wang2023comprehensive,van2022three}, also known as lifelong learning \cite{parisi2019continual} or incremental learning \cite{zhou2023deep}, offers an effective solution to these challenges. \highlight{It aims to develop systems capable of continuously learning and updating without forgetting past knowledge.} Recent advancements in CL methodologies have substantially enhanced the adaptability and knowledge retention capabilities of foundation LMs \cite{mehta2021empirical,cossu2022continual,lee2023pre}. 
\highlight{Notable successes have been documented in diverse downstream tasks, such as aspect-based sentiment analysis \cite{ke2021classic}, dialogue generation \cite{scialom2022fine}, text classification \cite{razdaibiedina2023progressive}, visual question answering \cite{zhang2023vqacl,qian2023decouple} and so on}. \highlight{Luo et al. \cite{luo2023empirical} conduct an empirical study on catastrophic forgetting (CF) in large language models (LLMs) during continual instruction tuning.} 
The aforementioned works underscore the potential of continual learning to significantly boost the performance of foundation LMs.


\tikzstyle{my-box}=[
    rectangle,
    draw=hidden-black,
    rounded corners,
    text opacity=1,
    minimum height=1.5em,
    minimum width=5em,
    inner sep=2pt,
    align=center,
    fill opacity=.5,
]
\tikzstyle{leaf}=[
    my-box, 
    minimum height=1.5em,
    fill=hidden-blue!90, 
    text=black,
    align=left,
    font=\normalsize,
    inner xsep=2pt,
    inner ysep=4pt,
]
\begin{figure*}[t]
    \vspace{-2mm}
    \centering
    \resizebox{\textwidth}{!}{
        \begin{forest}
            forked edges,
            for tree={
                grow=east,
                reversed=true,
                anchor=base west,
                parent anchor=east,
                child anchor=west,
                base=left,
                font=\large,
                rectangle,
                draw=hidden-black,
                rounded corners,
                align=left,
                minimum width=4em,
                edge+={darkgray, line width=1pt},
                s sep=3pt,
                inner xsep=2pt,
                inner ysep=3pt,
                line width=0.8pt,
                ver/.style={rotate=90, child anchor=north, parent anchor=south, anchor=center},
            },
            where level=1{text width=7.4em,font=\normalsize,}{},
            where level=2{text width=8.4em,font=\normalsize,}{},
            where level=3{text width=10.5em,font=\normalsize,}{},
            where level=4{text width=10.5em,font=\normalsize,}{},
            [
                Foundation LMs-based CL, ver
                [
                    Offline Continual \\ Learning ~(\S\ref{sect: Offline Continual Learning})
                    [
                        Domain-Incremental \\ Learning ~(\S\ref{sect: Domain-Incremental Learning})
                        [
                            PLMs-based DIL ~(\S\ref{sect: PLMs-based DIL})
                            [
                                LFPT5 \cite{qin2022lfpt}{,}
                                B-CL \cite{ke2021adapting}{,}
                                ELLE \cite{qin2022elle}{,}
                                AdapterCL \cite{madotto2020continual}{,} 
                                RMR\_ DSE \cite{li-etal-2022-overcoming}{,}
                                DEMIX \cite{geng2021continual} {,}
                                CLASSIC \cite{ke2021classic}{,} \\ 
                                Pretr \cite{cossu2022continual}{,}
                                CPT \cite{ke2022continualb}{,}
                                C-PT \cite{zhu2022continual}{,}
                                CL-KD \cite{castellucci-etal-2021-learning}{,}
                                PlugLM \cite{cheng2022language}{,}
                                AEWC \cite{lee2017toward}{,}
                                DAS \cite{ke2023continual}
                                , leaf, text width=44.6em
                            ]
                        ]
                        [
                            LLM-based DIL ~(\S\ref{sect: LLMs-based DIL})
                            [
                                COPF \cite{zhang2023copf}{,}
                                LAMOL \cite{sun2019lamol}{,}
                                RVAE\_LAMOL \cite{wang2022rvae}{,} 
                                Adapt-Retrieve-Revise \cite{zhang2023reformulating}{,}
                                Lifelong-MoE \cite{chen2023lifelong}{,} \\
                                DACP \cite{xie2023efficient}{,}
                                CPPO \cite{zhangcppo}{,}
                                EcomGPT-CT \cite{ma2023ecomgpt}{,} 
                                \highlight{LLM-CL \cite{ding2024boosting}{,}}
                                \highlight{AMA \cite{lin2024mitigatingalignmenttaxrlhf}}
                                , leaf, text width=44.6em
                            ]
                        ]
                        [
                            VLMs-based DIL ~(\S\ref{sect: VLMs-based DIL})
                            [
                                S-Prompt \cite{wang2022s}{,}
                                VQACL \cite{zhang2023vqacl}{,}
                                \highlight{SC-MLLM \cite{liu2024self}}{,}
                                \highlight{DIKI \cite{tang2024mind}}
                                , leaf, text width=44.6em
                            ]
                        ]
                    ]
                    [
                        Task-Incremental \\ Learning ~(\S\ref{sect: Task-Incremental Learning})
                        [
                            PLMs-based TIL ~(\S\ref{sect: PLMs-based TIL})
                            [
                                PP \cite{razdaibiedina2023progressive}{,}
                                CTR \cite{ke2021achieving}{,}
                                MeLL \cite{wang2021mell}{,}
                                LINC \cite{liu2021lifelong}{,}
                                ERDA \cite{qin2022continual}{,}
                                PCLL \cite{zhao2022prompt}{,}
                                BiHNet-Reg \cite{jin2021learn}{,}
                                ConTinTin \cite{yin2022contintin}{,} \\
                                HMI \cite{maekawa2023generative}{,} 
                                ACM \cite{zhang2022continual}{,}
                                DYNAINST \cite{mok-etal-2023-large}{,}
                                Conure \cite{10.1145/3404835.3462884}{,}
                                TERACON \cite{kim2023task}{,} \\
                                EMR \cite{wang-etal-2019-sentence}{,}
                                ERNIE 2.0 \cite{sun2020ernie}{,}
                                RecyclableTuning \cite{qin2023recyclable}{,}
                                \highlight{RecAdam \cite{chen2020recall}}
                                , leaf, text width=44.6em
                            ]
                        ]
                        [
                            LLMs-based TIL ~(\S\ref{sect: LLMs-based TIL})
                            [
                                Conpet \cite{song2023conpet}{,}
                                InstructAlign \cite{cahyawijaya2023instruct}{,}
                                Continual-T0 \cite{scialom2022continual}{,}
                                DynaMind \cite{du2023static}{,}
                                ELM \cite{jang2023exploring}{,}
                                O-LoRA \cite{wang-etal-2023-orthogonal}{,}
                                JARe \cite{peng2024scalable}{,} \\
                                \highlight{Robocoder \cite{li2024robocoder}}{,}
                                \highlight{Eureka \cite{ma2024eureka}}{,}
                                \highlight{COPF \cite{zhang2024copr}}
                                , leaf, text width=44.6em
                            ]
                        ]
                        [
                            VLMs-based TIL ~(\S\ref{sect: VLMs-based TIL})
                            [
                                Medical AI \cite{yi2023towards}{,}
                                CTP \cite{zhu2023ctp}{,}
                                ZSCL \cite{zheng2023preventing}{,}
                                MoE-Adapters4CL \cite{yu2024boosting}{,}
                                TRIPLET \cite{qian2023decouple}{,}
                                \highlight{AwoForget \cite{zhengadapt}}{,}\\
                                \highlight{SND \cite{yu2024select}}
                                , leaf, text width=44.6em
                            ]
                        ]
                    ]
                    [
                        Class-Incremental \\ Learning ~(\S\ref{sect: Class-Incremental Learning})
                        [
                            PLMs-based CIL ~(\S\ref{sect: PLMs-base CIL})
                            [
                                EPI \cite{wang2023rehearsal}{,}
                                IDBR \cite{huang2021continual}{,}
                                PAGeR \cite{varshney2022prompt}{,}
                                ENTAILMENT \cite{xia2021incremental}{,}
                                ExtendNER \cite{monaikul2021continual}{,}
                                PLE \cite{li2022continual}{,}
                                DE\&E \cite{wojcik2023domain}{,} \\
                                SRC \cite{liu2019continual}
                                , leaf, text width=44.6em
                            ]
                        ]
                        [
                            VLMs-based CIL ~(\S\ref{sect: VLMs-based CIL})
                            [
                                MoE-Adapters4CL \cite{yu2024boosting}{,}
                                VLM-PL \cite{kim2024vlm}{,}
                                Adaptation-CLIP \cite{liu2023class}{,}
                                PROOF \cite{zhou2023learning}{,}
                                LGCL \cite{khan2023introducing}{,}
                                ZSCL \cite{zheng2023preventing}{,} \\
                                CLAP \cite{jha2024clap4clip}{,}
                                GMM \cite{cao2024generative}{,}
                                \highlight{RAPF \cite{huang2024class}}{,}
                                \highlight{STAR-Prompt \cite{menabue2024semantic}}{,}
                                \highlight{DIKI \cite{tang2024mind}}{,}
                                \highlight{SND \cite{yu2024select}}{,}
                                \highlight{AwoForget \cite{zhengadapt}}
                                , leaf, text width=44.6em
                            ]
                        ]
                    ]
                ]
                [
                     Online Continual \\ Learning ~(\S\ref{sect: Online Continual Learning})
                    [
                        Hard Task \\ Boundary ~(\S\ref{sect: Hard Task Boundary})
                        [
                            PLMs-based HTB ~(\S\ref{sect: PLMs-based HTB})
                            [
                                MBPA++ \cite{de2019episodic}{,}
                                Meta-MBPA++ \cite{wang2020efficient}{,}  
                                OML-ER \cite{holla2020meta}{,}
                                TPEM \cite{geng2021continual}{,}
                                CID \cite{liu2021lifelong1}{,}
                                ProgModel \cite{shen2019progressive}
                                , leaf, text width=44.6em
                            ]
                        ]
                        [
                            VLMs-based HTB ~(\S\ref{sect: VLMs-based HTB})
                            [
                                \highlight{PEGP \cite{qiao2024gradient}}
                                , leaf, text width=44.6em
                            ]
                        ]
                    ]
                    [
                        Blurry Task \\ Boundary ~(\S\ref{sect: Blurry Task Boundary})
                        [
                            PLMs-based BTB ~(\S\ref{sect: PLMs-based BTB})
                            [
                                MBPA++ \cite{de2019episodic}{,}
                                Meta-MBPA++ \cite{wang2020efficient}{,}  
                                OML-ER \cite{holla2020meta}{,} 
                                TPEM \cite{geng2021continual}{,}
                                CID \cite{liu2021lifelong1}{,}
                                \highlight{S6 \cite{li2023online}}
                                , leaf, text width=44.6em
                            ]
                        ]
                        [
                            VLMs-based BTB ~(\S\ref{sect: VLMs-based BTB})
                            [
                                DKR \cite{cui2024continual}{,}
                                \highlight{SIT \cite{wang2024clip}{,}}
                                \highlight{OLiVia-Nav \cite{narasimhan2024olivia}}{,}
                                \highlight{G-NoCL \cite{seo2024just}}
                                , leaf, text width=44.6em
                            ]
                        ]
                    ]
                ]
            ]
        \end{forest}
    }
    \vspace{-6mm}
    \caption{Taxonomy of foundation language models for continual learning.}
    \label{fig:taxonomy}
    \vspace{-5mm}
\end{figure*}
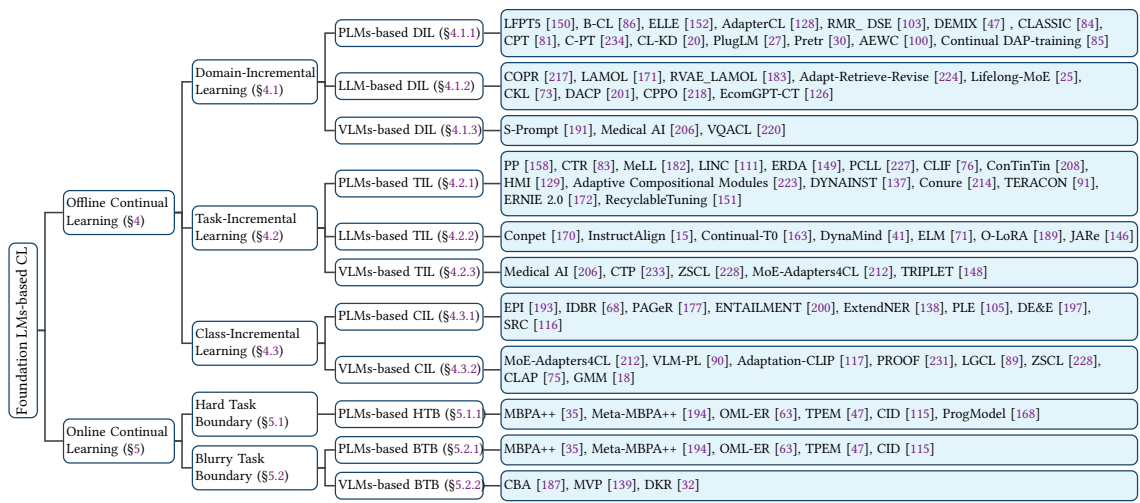

In the domain of continual learning, there has been a shift from traditional methods to those incorporating foundation LMs (Figure \ref{fig:IntroCLDifference}). First, foundation LMs has specialized transfer capability to quickly adapt to downstream tasks with only a few samples. Consequently, it is crucial to mitigate the degradation of both the zero-shot transfer and history task abilities while facilitating the acquisition of new skills. Second, due to the substantial number of parameters in foundation LMs, it is crucial to employ parameter-efficient techniques \cite{han2024parameter}, such as prompt tuning \cite{liu2022p} and adapters \cite{mundra2024comprehensive}, to update parameters without comprehensive retraining. Third, the foundation LMs possess the capability to follow instructions through instructional learning \cite{dong2022survey,ouyang2022training}, enabling more dynamic and context-aware interactions.

\highlight{This review systematically organizes continual learning strategies and technologies into two main categories: offline continual learning and online continual learning (Figure \ref{fig:taxonomy}). We begin by defining and explaining the different settings for these two types of continual learning. Offline continual learning includes domain/task/class-incremental CL, while online continual learning is further divided into methods that address hard task boundaries and those that manage blurry task boundaries \highlight{(Section \ref{sect: Settings and Learning Modes of CL})}. To further clarify the relationship between these strategies and model architectures, we group the methods based on three major model types: Pre-trained Language Models (PLMs), Large Language Models (LLMs), and Vision-Language Models (VLMs). These categories reflect the distinct requirements for continual learning in each architecture and emphasize the role these models play in shaping learning strategies. Additionally, we classify the methods into four key approaches: traditional continual learning methods, continual pre-training methods, parameter-efficient tuning methods, and instruction-based methods \highlight{(Section \ref{sect: Offline Continual Learning}, \ref{sect: Online Continual Learning})}. This categorization highlights the different techniques used to address challenges like catastrophic forgetting and to improve knowledge transfer in continual learning scenarios}. Finally, we static the main datasets from various perspectives \highlight{(Section\ref{sect: Datasets})} and review the key metrics to evaluate the forgetting and transferring of the models \highlight{(Section \ref{sect: Metrics})}.

The main contributions of this survey paper can be summarized as follows:
\begin{itemize}[leftmargin=*, align=left]
    \item We thoroughly review the existing literature on foundation LMs-based CL approaches, which integrate foundation LMs with CL to learn new knowledge without retraining the models. It is quite different from traditional CL since foundation LMs have great abilities of transfer learning, zero-shot and instruction following with huge parameters.
    \item We give the definitions of different settings and categorize these studies into various classes to better understand the development of this domain. In addition to the traditional methods like replay, regularization and parameter-isolation-based algorithms, we also summarize the works about continual pre-training methods, parameter-efficient tuning methods and instruction tuning-based methods. 
    \item We provide the characters of existing datasets for CL and present the main metrics to evaluate the performance of preventing forgetting and knowledge transfer. Furthermore, we discuss the most challenging problems of foundation LMs-based CL and point out promising future research directions in this field.
\end{itemize}

\section{Related Surveys} 
\label{sect: Related Surveys}

\subsection{Continual Learning}
Early examinations have provided broad coverage, as observed in surveys such as Parisi et al. \cite{parisi2019continual}. Recently, Wang et al. \cite{wang2023comprehensive} conduct a comprehensive survey that categorizes five key strategies in CL: regularization-based, replay-based, optimization-based, representation-based, and architecture-based approaches. This survey reflects an effort to organize and understand the diverse methodologies employed in the field. Notably, there is also a growing focus on class-incremental setting \cite{zhou2023deep,masana2022class,belouadah2021comprehensive} and replay-based approaches \cite{hayes2021replay}.
\highlight{Recent surveys \cite{wu2024continual, shi2024continual, zheng2024towards} provide comprehensive overviews of continual pre-training, continual domain-adaptive pre-training, continual instruction tuning, and continual alignment. In particular, Shi et al. \cite{shi2024continual} provides an in-depth overview of CL, specifically in LLMs, introducing vertical CL (adapting from general to specific domains) and horizontal CL (adapting across time and domains). Additionally, one notable approach in continual alignment is Reinforcement Learning from Human Feedback (RLHF) \cite{wu2022survey}, which incorporates human feedback to optimize model outputs for better task adaptation.}

\subsection{Continual Learning for Computer Vision}
In the realm of computer vision (CV), De et al. \cite{de2021continual} concentrate on task incremental classification and offer significant contributions including a taxonomy, a new framework for balancing stability and plasticity, and an extensive comparison of continual learning methods against baselines.
Qu et al. \cite{qu2021recent} present a comprehensive examination of continual learning, highlighting its vital role in the accumulation of knowledge from sequential data streams. This research explores multiple methods including regularization, knowledge distillation, memory-based approaches, and more, categorizing them by their characteristics and applications in CV.
Moreover, Mai et al. \cite{mai2022online} focus on the realm of online continual learning within the image classification task. This study evaluates the efficacy of state-of-the-art methods across diverse memory and data configurations. 
Masana et al. \cite{masana2022class} conduct a comprehensive evaluation of class-incremental methods for image classification, involving large-scale datasets and various network architectures. Belouadah et al. \cite{belouadah2021comprehensive} focus more on class-incremental learning algorithms specifically for visual tasks, defining key properties of these algorithms, formalizing the class-incremental learning problem, and providing an evaluation framework for thorough analysis.

\subsection{Continual Learning for NLP}
Biesialska et al. \cite{biesialska2020continual} address the challenge of continual learning within Natural Language Processing (NLP), wherein conventional architectures struggle to accommodate new tasks without compromising previously \highlight{acquired knowledge. 
In a similar vein}, Ke et al. \cite{ke2022continualc} offer a focused survey on continual learning within the NLP domain, providing a comprehensive examination of various continual learning settings, methodologies, and challenges.
This work presents an in-depth analysis of state-of-the-art approaches and extends original CL settings to be more general \highlight{and up-to-date. }
\highlight{Gogoulou et al. \cite{gogoulou2023study} study the pros and cons of updating a language model when new data comes from new languages – the case of CL under language shift. They feed various languages into the model to examine the impact of pre-training sequence and linguistic characteristics on both forward and backward transfer effects across three distinct model sizes.
}
\subsection{Continual Learning for Other Domains}

Recent surveys, surveys like \cite{zhang2023survey, shaheen2022continual,lesort2020continual} explore advancements in incremental learning across other domains. Zhang et al. \cite{zhang2023survey} focus on neural recommendation systems, particularly introducing the Incremental Update Recommendation Systems (IURS) to bridge the gap between academic research and industrial applications. They discuss the unique challenges of IURS compared to traditional Batch Update Recommendation Systems (BURS) and offer a detailed review of existing literature and evaluation methods. Shaheen et al. \cite{shaheen2022continual} offer a comprehensive overview of the CL methods in real-world contexts, highlighting efficient learning algorithms for handling large sequential datasets within resource constraints and exploring the applicability of CL to autonomous systems. Lesort et al. \cite{lesort2020continual} investigate CL in robotics, defining it as a dynamic learning paradigm where both data distribution and objectives evolve. They also emphasize the challenges in evaluating CL algorithms and introduce a new framework with tailored metrics for assessing CL methods.

\highlight{
Continual learning across domains aims to mitigate catastrophic forgetting by integrating new tasks while preserving knowledge from prior ones \cite{monaikul2021continual,li2022continual}. However, domains like NLP, CV, and robotics exhibit significant differences. CV methods \cite{wang2022dualprompt, wang2022learning} typically focus on classification, while NLP involves more complex tasks such as information extraction and text generation. Robotics tasks evolve sequentially, relying on decision-making and state transitions \cite{kirkpatrick2017overcoming,aljundi2018memory}. NLP models, primarily based on PLMs or LLMs, excel in few-shot and transfer learning, whereas CV tasks commonly utilize traditional neural networks (e.g., ResNet). Recently, Vision-Language Models (VLMs) have been introduced in CV to enhance continual learning. Additionally, the pre-trained transformers in NLP enables parameter-efficient strategies like adapter and prompt tuning, allowing adaptation to new tasks with minimal retraining \cite{ke2022continualb,madotto2020continual}.}

This paper centers on the advancements in CL as applied to foundation LMs, which have obtained success in the fields of NLP and multimodal. We categorize existing works into offline and online CL based on PLMs, LLMs, and VLMs.

\begin{figure*}[t!]
\vspace{-2mm}
    \centering
    \includegraphics[width=0.9\linewidth]{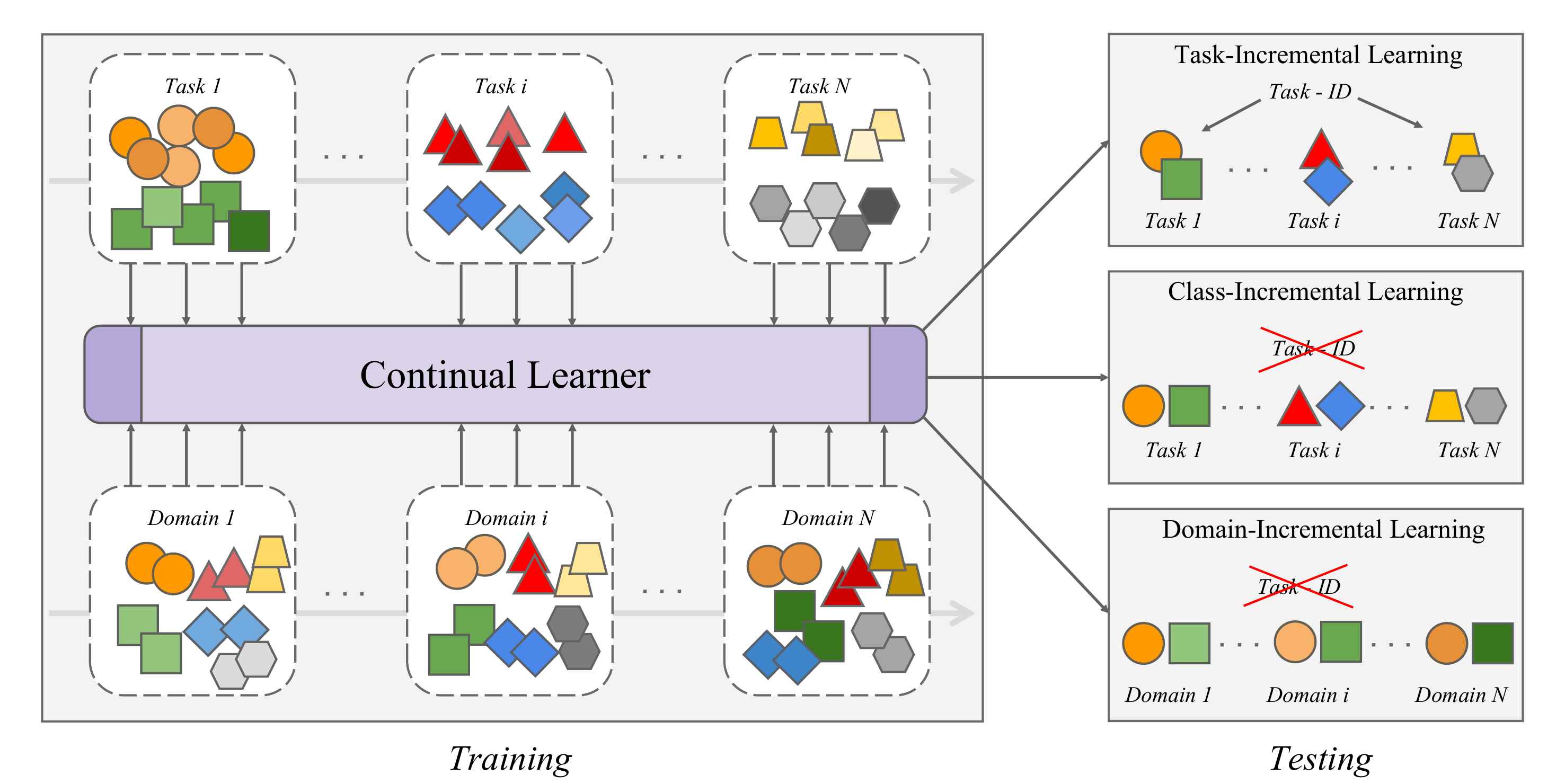}
    \vspace{-4mm}
    \caption{The setting of different offline continual learning tasks, including task-incremental learning, class-incremental learning and domain-incremental learning. The samples with different classes (domains) are marked with various shapes (colors).}
    \label{fig:offline tasks}
    \vspace{-3mm}
\end{figure*}

\begin{figure*}[t!]
    \centering
    \includegraphics[width=0.9\linewidth]{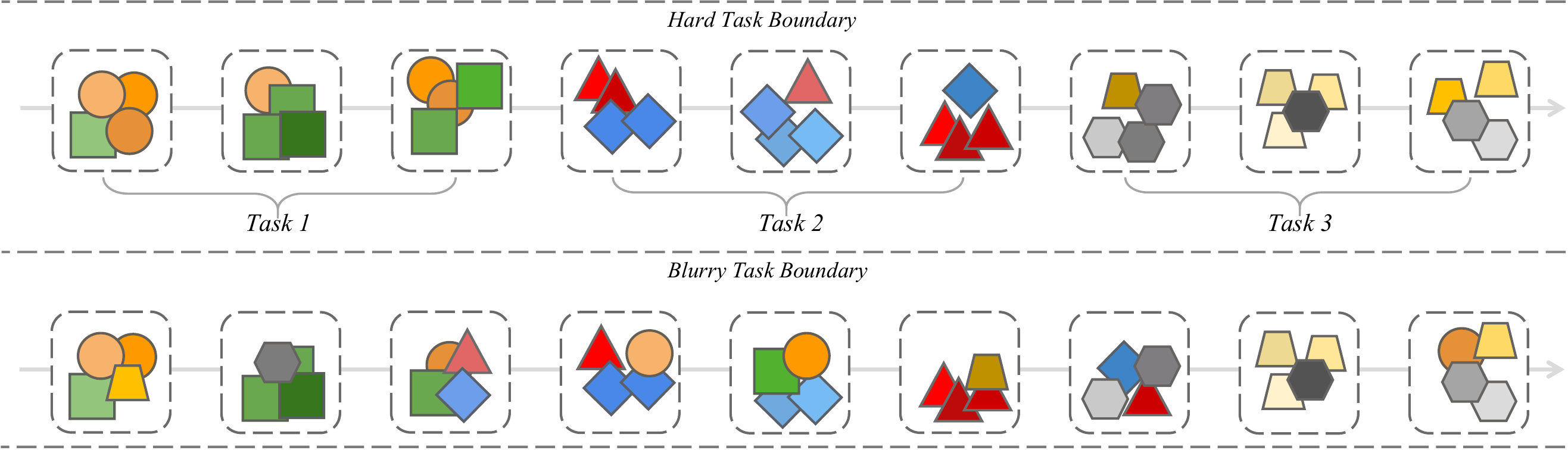}
    \vspace{-3mm}
    \caption{The setting of different online continual learning tasks, including hard task boundary arriving and blurry task boundary arriving. The samples with different classes (domains) are marked with various shapes (colors).}
    \label{fig:online tasks}
    \vspace{-4mm}
\end{figure*}

\section{Typical Foundation Language Model} 
\paragraph{\highlight{Pre-trained Language Models (PLMs).}} 
\highlight{PLMs are trained on large-scale textual data using techniques such as Masked Language Modeling (MLM) and Next Sentence Prediction (NSP) for various tasks, including text classification, sentiment analysis, and named entity recognition. Notable examples of PLMs include ELMo \cite{sarzynska2021detecting}, BERT \cite{devlin2018bert}, early versions of the GPT series \cite{radford2018improving}, and RoBERTa \cite{liu2019roberta}.
PLM-based methods aim to mitigate catastrophic forgetting when learning new tasks or domains within a closed-world setting. Continual learning for PLMs focuses on parameter-efficient tuning and replay-based strategies to reduce resource consumption and prevent catastrophic forgetting. Approaches like adapter-based tuning, prompt-based methods, and knowledge distillation are popular because they update the model with few parameters. This ensures that PLMs remain feasible to deploy on standard hardware while keeping computational costs low.
}
\paragraph{\highlight{Large Language Models (LLMs).}} 
\highlight{LLMs are significantly larger in scale than PLMs, often containing billions or even trillions of parameters. These models are trained on vast datasets using language modeling techniques and excel in tasks such as generation, instruction learning, and in-context learning. LLMs are typically exposed to a broader range of complex tasks, from domain-specific corpora to open-domain challenges.
Prominent examples include OpenAI's GPT-3 \cite{brown2020language}, GPT-4 \cite{achiam2023gpt}, LLaMA \cite{touvron2023llama}, and others. Due to the substantial memory and computational demands associated with LLMs, continual learning strategies for LLMs often focus on optimizing memory efficiency (e.g., orthogonal low-rank adaptation, recyclable tuning) and selectively updating parameters (e.g., Mixture-of-Experts, dynamic task selection), as retraining or fine-tuning the entire model is typically cost-prohibitive. Additionally, LLMs require techniques to mitigate catastrophic forgetting while managing task diversity and retaining foundation knowledge.
}
\paragraph{\highlight{Vision-Language Models (VLMs).}} 
\highlight{VLMs are designed to process and integrate multimodal information, such as text and vision, by building on PLMs or LLMs for tasks like image captioning, visual question answering, and visual reasoning. 
Notable examples include CLIP \cite{radford2021learning} and ALIGN \cite{jia2021scaling}. 
VLMs must effectively handle multimodal data, which necessitates continual learning methods that can efficiently manage cross-modal learning. Continual learning for VLMs often emphasizes alignment across multiple modalities using specialized modules such as cross-modal fusion techniques, attention mechanisms, and lightweight adapters. Additionally, methods that focus on retaining modality-specific knowledge and enabling parameter-efficient fine-tuning (e.g., Adaptation-CLIP \cite{liu2023class}, MoE-Adapters4CL \cite{yu2024boosting}) are essential due to the high costs associated with continual learning across both vision and language domains.
}

\section{Settings and Learning Modes of CL}
\label{sect: Settings and Learning Modes of CL}

\subsection{Basic Formulation}
\label{sect: Basic Formulation}
Continual learning is an advanced method in machine learning. Within this framework, the model is sequentially trained across a diverse array of tasks denoted as \(t\) within the set \(T = \{1, 2, ..., N\}\), where each task \(t\) is associated with its individual dataset \(X_t = \{(x^{(t)}_i, y^{(t)}_i)\}^{|X_t|}_{i=1}\). Here, \(x^{(t)}_i\) represents an individual training example, and \(y^{(t)}_i\) denotes the corresponding class label for task \(t\), while \(|X_t|\) indicates the total number of samples in task \(t\). However, the data distributions between any two tasks \(t\) and \(t^\prime\) are distinct (\(p(X_t) \neq p(X_{t^\prime})\) for all \(t \neq t^\prime\)). This distinction presents a fundamental challenge in managing the diversity of data distributions across multiple tasks. This setup necessitates the model to learn new knowledge while retaining past information.

Continual learning encompasses two principal paradigms: offline and online continual learning. These paradigms define how data arrives and how the model updates its knowledge over time.
\begin{itemize}[leftmargin=*, align=left]
    \item Offline Continual Learning: This setting involves learning across a series of tasks, with each task fully presented before handling the next task. For each task \( t \), the model trains on the entire dataset \( D_t \) through multiple epochs. The model progresses to task \( t+1 \) only upon achieving the desired proficiency on task \( t \). 
    \item Online Continual Learning: This setting operates within a dynamic framework wherein the model learns knowledge from a stream of data points or mini-batches presented sequentially. Additionally, the model lacks access to the entire dataset for a given task. This setting closely mirrors real-world scenarios characterized by continuous data flow, compelling the model to adapt in real time.

\end{itemize}

\subsection{Typical Scenarios}
\label{sect: Typical Scenario}
\subsubsection{Offline Continual Learning}
Offline CL (Figure \ref{fig:offline tasks}) comprises three principal scenarios, each distinguished by distinct characteristics: Domain-Incremental Learning \highlight{(e.g., LFPT5 \cite{qin2022lfpt}, LAMOL \cite{sun2019lamol}, S-Prompt \cite{wang2022s})}, Task-Incremental Learning \highlight{(e.g., PP \cite{razdaibiedina2023progressive}, Conpet \cite{song2023conpet}, ZSCL \cite{zheng2023preventing})}, and Class-Incremental Learning \highlight{(e.g., EPI \cite{wang2023rehearsal}, IDBR \cite{huang2021continual}, GMM \cite{cao2024generative})}.

\begin{itemize}[leftmargin=*, align=left]
    \item Domain-Incremental Learning (DIL): The model aims to process diverse data distributions. Specifically, in DIL, while the data distributions \( p(X_t) \) in task \( t \) and \( p(X_{t'}) \) in task \( t' \) are different, their task types and class labels remain consistent. The task identities (task IDs) are not required.
    \item Task-Incremental Learning (TIL): The model is designed to handle a series of tasks, each with unique objectives. The classes within these tasks may or may not be disjoint. The boundaries of each task are clear, and task IDs are provided during both the training and testing phases.
    \item Class-Incremental Learning (CIL): The model is designed to continually learn new class information while retaining knowledge of previously learned classes. For tasks \( t \) and \( t' \), while they might share the same task type (such as classification), their class sets \( C_t \) and \( C_{t'} \) are distinct. Moreover, the task IDs are only available during training.
\end{itemize}

In summary, DIL concentrates on adapting the model to the shifts in input data distributions while maintaining consistency in tasks and classes. TIL necessitates the model's ability to learn and retain task-specific knowledge over successive tasks. On the other hand, CIL highlights the gradual integration of new classes into the model's recognition capabilities without compromising knowledge of previously learned classes.

\subsubsection{Online Continual Learning}
In online continual learning (Figure \ref{fig:online tasks}), the existing researches are categorized into two configurations based on the arrival pattern of tasks: "Hard Task Boundary" \highlight{(e.g., MBPA++ \cite{de2019episodic}, PEGP \cite{qiao2024gradient})} and "Blurry Task Boundary" \highlight{(e.g., SIT \cite{wang2024clip}, G-NoCL \cite{seo2024just})}:

\begin{itemize}[leftmargin=*, align=left]
    \item Hard Task Boundary: The arrival of tasks follows a strictly structured and sequential process. Data from the preceding task is completely processed before transitioning to the next task, ensuring no overlap of data between tasks.
    \item Blurry Task Boundary: The distinction between tasks is less clear, similar to real-world scenarios. Data from different tasks are intermixed, making it difficult to pinpoint when one task ends and another begins.
\end{itemize}

In both setups, the main challenge lies in achieving the balance of learning new data while preserving previously gained knowledge, often termed as catastrophic forgetting. Numerous approaches, such as experience replay \cite{qin2022elle,sun2019lamol}, elastic weight consolidation (EWC) \cite{kirkpatrick2017overcoming}, and progressive neural networks \cite{ke2021adapting,ke2021achieving}, have emerged to address this issue. Each method comes with its unique strengths and weaknesses upon the task arrival configuration.

\section{Offline Continual Learning}
\label{sect: Offline Continual Learning}

\subsection{Domain-Incremental Learning}
\label{sect: Domain-Incremental Learning}

\subsubsection{PLMs-based DIL}
\label{sect: PLMs-based DIL} 
\begin{figure}[t!]
\vspace{-2mm}
    \centering
    \begin{subfigure}[b]{0.37\textwidth}
        \centering
        \includegraphics[width=\textwidth]{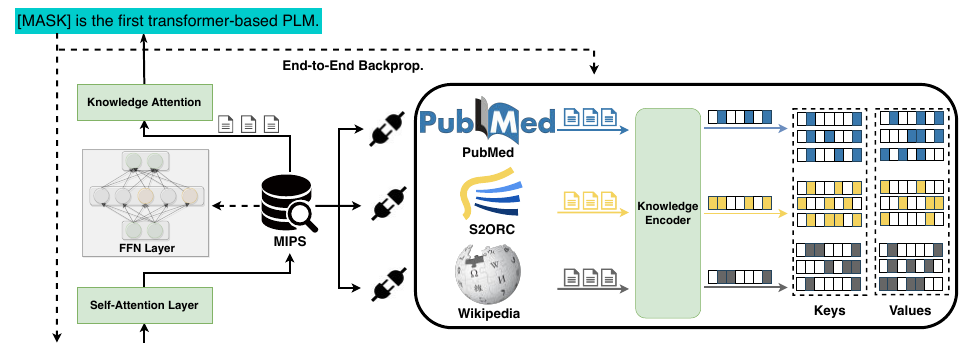}
        \caption{PlugLM}
        \label{fig:pluglm}
    \end{subfigure}
    \hfill
    \begin{subfigure}[b]{0.25\textwidth}
        \centering
        \includegraphics[width=\textwidth]{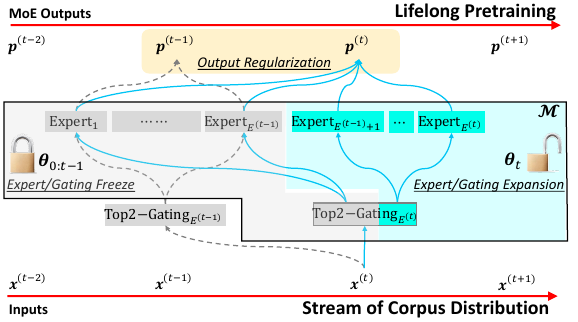}
        \caption{Lifelong-MoE}
        \label{fig:Lifelong-MoE}
    \end{subfigure}
    \hfill
    \begin{subfigure}[b]{0.28\textwidth}
        \centering
        \includegraphics[width=\textwidth]{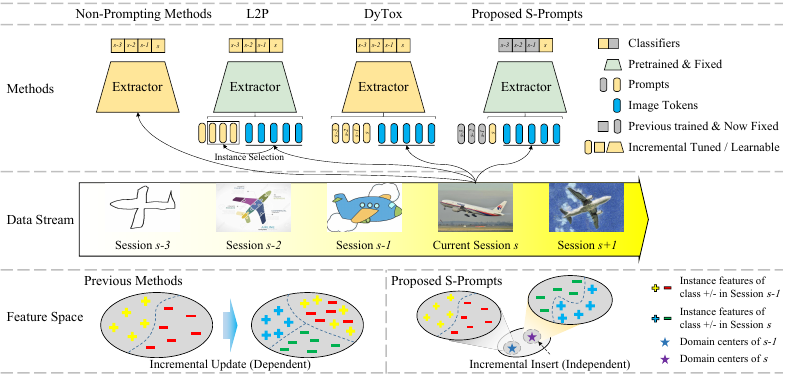}
        \caption{S-Prompts}
        \label{fig:S-Prompts}
    \end{subfigure}
    \vspace{-3mm}
    \caption{Frameworks in DIL: PlugLM (PLM-based) \cite{cheng2022language}, Lifelong-MoE (LLM-based) \cite{chen2023lifelong}, S-Prompts (VLM-based) \cite{wang2022s}.}
    \label{fig:main1}
        \vspace{-5mm}
\end{figure}

\paragraph{Traditional Methods.} Continual Learning methods are increasingly used with Pre-trained Language Models (PLMs), adopting various techniques such as replay, regularization, and parameter-isolation. Driven by regularization, RMR\_DSE \cite{li-etal-2022-overcoming} optimizes recall and uses domain drift estimation to manage embedding space shifts across domains\highlight{, effectively reducing catastrophic forgetting in seq2seq tasks under memory constraints.} CL-KD \cite{castellucci-etal-2021-learning} uses a Teacher-Student model to transfer language knowledge efficiently. AEWC \cite{lee2017toward} is an improved EWC algorithm, where the importance can be computed in an online manner. DEMIX \cite{gururangan2021demix} is a parameter-isolation approach, utilizing a collection of domain-specific experts for dynamic domain adaptation. \highlight{It offers superior performance in both in-domain and out-of-domain tasks, providing strong generalization to unseen domains and promising improved adaptability and scalability in multi-domain settings.} Lastly, PlugLM \cite{cheng2022language}  (Figure \ref{fig:pluglm}) features a differentiable plug-in memory (DPM) for domain-adaptive training\highlight{, with a flexible key-value structure that decouples knowledge storage from model parameters, enabling more effective knowledge management in domain-adaptive training.}

\paragraph{Continual Pre-training Methods.}
Pretr \cite{cossu2022continual} formalizes continual pre-training, where models are continuously pre-trained on sequential data streams before fine-tuning for downstream tasks. \highlight{DAS \cite{ke2023continual} offers an effective solution for continual domain-adaptive pre-training (DAP-training) of LMs. By integrating soft-masking, KL-divergence loss, and contrastive learning, it mitigates catastrophic forgetting and enhances knowledge transfer between domains.}

\paragraph{Parameter-Efficient Tuning Methods.} Due to the huge parameters of LMs, parameter-efficient tuning methods like adaptors \cite{houlsby2019parameter,poth2023adapters} and p-tuning \cite{liu2022p} are used for domain-incremental CL \cite{madotto2020continual,ke2021adapting,ke2021classic,zhu2022continual}.

The adapter architecture incorporates a skip-connection to minimize the number of parameters. A notable exemplar of this approach is AdapterCL \cite{madotto2020continual}, which employs residual adapters tailored for task-oriented dialogue systems. \highlight{This makes it ideal for real-world deployment with frequent domain updates without the need for full retraining.} In a related vein, B-CL \cite{ke2021adapting} \highlight{sets a new standard in Aspect-Based Sentiment Classification} by utilizing CL adapters within capsule network architectures\highlight{, demonstrating that backward knowledge transfer significantly boosts performance while preserving task-specific information.}
CLASSIC \cite{ke2021classic} addresses catastrophic forgetting by leveraging adapters in BERT\highlight{, employing a contrastive CL strategy that facilitates knowledge transfer across tasks without requiring task identifiers during testing.}
Furthermore, CPT \cite{ke2022continualb} incorporates CL-plugins in RoBERTa\highlight{, effectively improving end-task performance by managing domain-specific knowledge transfer, particularly in scenarios with limited labeled data.}

Prompt tuning \cite{liu2022p}, or P-tuning, introduces trainable prompts into the sequence of input word embeddings, while the LM remains frozen. LFPT5 \cite{qin2022lfpt} integrates prompt tuning, pseudo-labeled samples, and KL divergence to excel in lifelong few-shot language learning (LFLL). \highlight{The use of pseudo-labeled samples and KL divergence helps maintain label consistency between models, though task transferability requires careful attention}. Similarly, C-PT \cite{zhu2022continual} enhances knowledge transfer between tasks in dialogue systems through prompt initialization, query fusion, and memory replay\highlight{, outperforming baselines like EWC, Replay, and AdapterCL, while being efficient in parameter usage and memory.}

\paragraph{Instruction Tuning-based Methods.}
Instruction tuning-based methods involve transforming a given task into natural language instructions. ELLE \cite{qin2022elle} incorporates expanding streaming data into PLMs with two key components: (1) function-preserved model expansion, which enhances knowledge acquisition efficiency by changing the width and depth of PLM, and (2) pre-trained domain prompts, which enhance downstream task adaptation by \highlight{utilizing domain-specific knowledge from the pre-training phase.}

\subsubsection{LLMs-based DIL}
\label{sect: LLMs-based DIL}

\paragraph{Traditional Methods.}

\highlight{In Lifelong Language Learning (LLL), a key challenge is training models on sequential NLP tasks while preserving knowledge from previous tasks.} LAMOL \cite{sun2019lamol} tackles this by generating pseudo-samples from past tasks during new task training, effectively reducing knowledge loss without additional memory or computational overhead. RVAE\_LAMOL \cite{wang2022rvae} builds on this by utilizing a variational autoencoder (RVAE) to map tasks into a unified semantic space, while introducing Alternate Lag Training (ALT) and an identity task \highlight{to improve stability and task-specific sample generation}.

\highlight{Efficiently and stably aligning with dynamic human preferences is challenging, especially when minimizing retraining costs.} CPPO \cite{zhangcppo} addresses this issue by integrating sample-wise weighting into the PPO algorithm, effectively balancing policy learning and knowledge retention\highlight{, particularly in adaptive tasks like summarization and question-answering.} Similarly, COPF \cite{zhang2023copf} reduces reliance on human feedback by regularizing the current policy with past optimal policies. \highlight{It is well-suited for handling unlabeled data, making it ideal for real-world applications with limited labeled datasets. Furthermore, AMA \cite{lin2024mitigatingalignmenttaxrlhf} addresses the "alignment tax"—the trade-off between maximizing alignment rewards and mitigating forgetting—by interpolating between pre- and post-RLHF model weights. This interpolation identifies optimal layer-wise weight combinations, balancing "alignment tax" across various RLHF algorithms.}


\paragraph{Continual Pre-training Methods} 

Large language models (LLMs) excel in open-domain tasks but face significant challenges in domain-specific applications, such as insufficient domain knowledge, limited capacity to leverage that knowledge, and inadequate adaptation to specialized data formats. To overcome these limitations, continual pre-training \cite{xie2023efficient,ma2023ecomgpt,cheng2023adapting} has emerged as a promising approach, enabling LLMs to better fit domain-specific tasks. For example, AdaptLLM \cite{cheng2023adapting} transforms raw corpora into reading comprehension texts\highlight{, but it was found that while pre-training on raw data enhances domain knowledge, it hampers the model’s ability to answer questions effectively}. As an alternative to full retraining, Domain-Adaptive Continual Pre-training (DACP) leverages large domain-specific corpora to continually train models, although it involves high computational costs. DACP \cite{xie2023efficient} introduces Efficient Task-Specific (ETS-DACP) and Task-Agnostic (ETA-DACP) strategies to \highlight{optimize performance by focusing either on task-specific foundational models or by selecting the most informative domain samples}. \highlight{Given the high cost of training from scratch and limited annotated data}, EcomGPT-CT \cite{ma2023ecomgpt} \highlight{explores continual pre-training on unlabeled general and E-commerce corpora,} employing a data-mixing strategy to better handle semi-structured data. Experimental results across multiple tasks show \highlight{continual pre-training and data-mixing strategies improve few-shot and zero-shot performance}.

\paragraph{Parameter-Efficient Tuning Methods.}
Lifelong-MoE \cite{chen2023lifelong} (Figure \ref{fig:Lifelong-MoE}) uses a Mixture-of-Experts (MoE) architecture, expanding capacity by adding new experts while freezing prior experts and gating mechanisms. \highlight{It surpasses existing methods with better few-shot performance on 19 NLP tasks.}

\paragraph{Instruction Tuning-based Methods.}
\highlight{Adapt-Retrieve-Revise (ARR)\cite{zhang2023reformulating} aims at reducing hallucinations in specialized domains such as the Chinese legal domain. It consists of three steps: adapting a 7-billion-parameter language model for initial responses, retrieving corroborative evidence from an external knowledge base, and integrating these to refine the final response with GPT-4.
Furthermore, LLM-CL \cite{ding2024boosting}, designed for aspect-based sentiment analysis, employs task-specific prompts and independently learns both domain-shared and domain-specific knowledge.}

\subsubsection{VLMs-based DIL}
\label{sect: VLMs-based DIL}

Vision-language models (VLMs) excel in domain-incremental learning, particularly through methods like S-Prompt \cite{wang2022s} (Figure \ref{fig:S-Prompts}), which independently learns prompts across diverse domains using pre-trained VLMs. This approach introduces innovative image and language-image prompt acquisition techniques, utilizing a unified cross-entropy loss during training and a k-nearest neighbors (K-NN) method to identify domains during inference. In visual question answering, VQACL \cite{zhang2023vqacl} enhances multimodal tasks through a dual-level task sequence. This framework uses a compositionality test to assess generalization to new skill-concept combinations and employs a representation learning strategy that differentiates sample-specific (SS) features, capturing unique input attributes, from sample-invariant (SI) features, which retain essential characteristics. \highlight{DIKI \cite{tang2024mind} preserves pre-trained knowledge by using a residual attention mechanism to inject new information into a frozen backbone, minimizing interference. A distribution-aware calibration scheme ensures smooth integration of new knowledge while maintaining the VLM's zero-shot capabilities. SC-MLLM \cite{liu2024self} is a robotic manipulation framework that utilizes an exponential moving average (EMA) for continuous learning, reducing the risk of catastrophic forgetting.}

\subsection{Task-Incremental Learning}
\label{sect: Task-Incremental Learning}

\subsubsection{PLMs-based TIL}
\label{sect: PLMs-based TIL}
\paragraph{Traditional Methods.} 

\begin{figure}[t!]
\vspace{-2mm}
    \centering
    \begin{subfigure}[b]{0.3\textwidth}
        \centering
        \includegraphics[width=\textwidth]{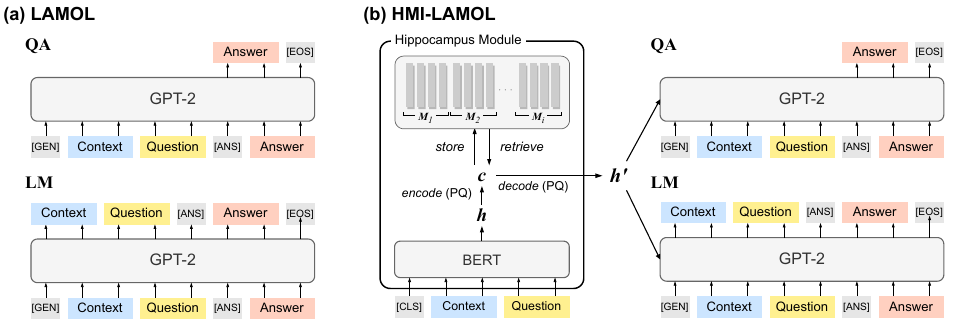}
        \caption{HMI}
        \label{fig:hmi}
    \end{subfigure}
    \hfill
    \begin{subfigure}[b]{0.3\textwidth}
        \centering
        \includegraphics[width=\textwidth]{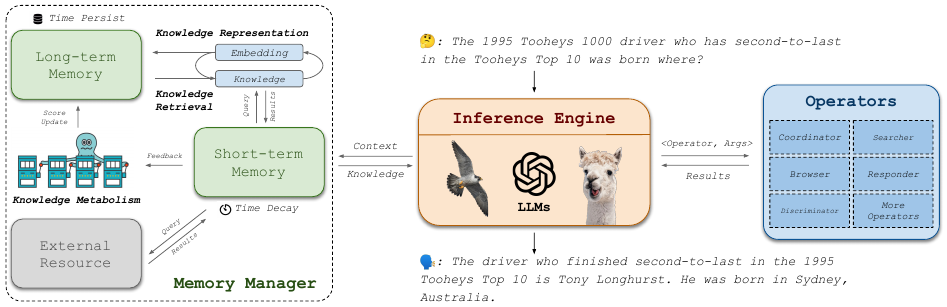}
        \caption{DynaMind}
        \label{fig:dynamind}
    \end{subfigure}
    \hfill
    \begin{subfigure}[b]{0.3\textwidth}
        \centering
        \includegraphics[width=\textwidth]{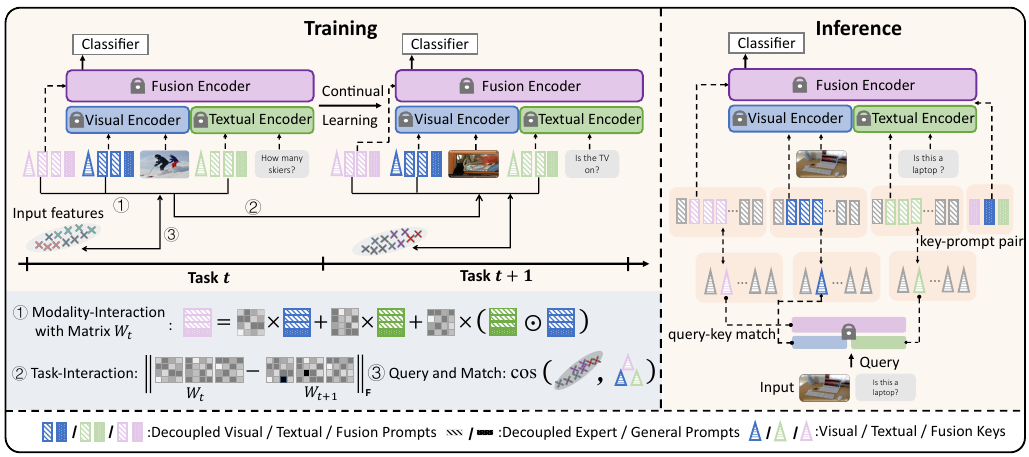}
        \caption{TRIPLET}
        \label{fig:triplet}
    \end{subfigure}
        \vspace{-3mm}
    \caption{Frameworks in TIL: HMI (PLM-based) \cite{maekawa2023generative}, DynaMind (LLM-based) \cite{du2023static}, TRIPLET (VLM-based) \cite{qian2023decouple}.}
    \label{fig:main2}
        \vspace{-6mm}
\end{figure}

Drawing inspiration from neurobiological mechanisms, HMI \cite{maekawa2023generative} (Figure \ref{fig:hmi}) integrates compressed representations of prior training instances\highlight{, providing selective guidance for generating training samples}. 
\highlight{Unlike traditional Continual Relation Learning (CRL) that rely on large labeled datasets. Continual Few-Shot Relation Learning (CFRL) focuses on learning new relational patterns with minimal labeled data.} ERDA \cite{qin2022continual} tackles this challenge by selecting informative samples from unlabeled data and enforcing relational constraints in the embedding space. \highlight{As for Lifelong Relation Extraction,} EMR \cite{wang-etal-2019-sentence} utilizes a working memory mechanism to selectively replay stored samples\highlight{, helping the model retain prior knowledge while learning new tasks.}

\highlight{Learning user representations is fundamental for personalized recommender systems, but current approaches often train separate models for each task, which leads to inefficiency.} To improve this, Conure \cite{10.1145/3404835.3462884} introduces a framework that manages multiple tasks by pruning less critical parameters to accommodate new, task-specific ones. This method allows for positive transfer learning and preserves essential parameters to avoid knowledge loss. However, learning task-specific user representations for each task is impractical. \highlight{Recent studies focus on universal user representations, which capture generalized user traits relevant to multiple tasks.} TERACON \cite{kim2023task} addresses this challenge by employing task-specific soft masks to isolate parameters and ensure effective knowledge retention.

\highlight{Given that PLMs often suffers catastrophic forgetting when applied to CL scenarios,} CTR \cite{ke2021achieving} tackles this by utilizing innovative techniques such as CL-plugins and task masking, which significantly improve knowledge transfer while reducing forgetting. \highlight{This approach proves particularly effective on challenging benchmarks, such as the 20News dataset, where task dissimilarity often causes severe forgetting in other models. RecAdam \cite{chen2020recall} extends the concepts of EWC by applying a regularization term to the pre-trained weights, utilizing an annealing coefficient to progressively incorporate the importance of prior knowledge.}

The diversity of text distributions and intents across domains poses challenges for scaling user intent detection (UIC) in industrial applications. MeLL \cite{wang2021mell} proposes a solution by leveraging global and local memory networks to maintain cross-task representations\highlight{, which allows the model to continually adapt to new tasks without significant loss of performance.} MeLL also employs an LRU policy for efficient memory management and controls parameter growth\highlight{, making it scalable for real-world applications.}

Recent advancements in conversational AI focus on overcoming the limitations of traditional chatbots, which rely on static knowledge bases and extensive manual data annotation. Lifelong INteractive learning in Conversation (LINC) introduces a dynamic learning framework \highlight{that mirrors human cognitive processes, enabling chatbots to integrate and utilize knowledge in real-time} \cite{liu2021lifelong,mazumder2024lifelong,liu2020learning}. Structured with an Interaction Module, Task Learner, and Knowledge Store, LINC enhances chatbots by improving real-time information extraction, handling erroneous inputs, and refining conversational skills, thereby boosting linguistic and interactive capabilities.

\paragraph{Continual Pre-training Methods.}
Continual pre-training represents a paradigm where PLMs are progressively enhanced by assimilating new knowledge from expanding datasets. ERNIE 2.0 \cite{sun2020ernie} \highlight{illustrates this approach by incrementally} constructing pre-training tasks, enabling the model to capture increasingly complex lexical, syntactic, and semantic nuances. Unlike traditional fixed-task training, it employs a continual multi-task learning framework to enhance its capabilities. \highlight{Advancing this field further,} RecyclableTuning \cite{qin2023recyclable} introduces two strategies: initialization-based and distillation-based. The former builds on fine-tuned weights from existing PLMs, while the latter reuses outdated weights to preserve knowledge and improve efficiency in future models.

\paragraph{Parameter-Efficient Tuning Methods.}
Expounding upon the crucial need for more efficacious knowledge integration, ACM \cite{zhang2022continual} dynamically adjusts transformer architectures and leverages pseudo-experience replay to enhance knowledge transfer. Continual Learning of Few-Shot Learners (CLIF) addresses a gap between \highlight{rapid generalization methods (such as few-shot learning) and traditional continual learning approaches, which are not inherently designed for such generalization. BiHNet-Reg \cite{jin2021learn} provides a solution by generating task-specific adapter weights through bi-level task representations and incorporating regularization.}

\paragraph{Instruction Tuning-based Methods.}

PCLL \cite{zhao2022prompt}, designed for task-oriented dialogue systems, uses a conditional variational autoencoder with prompts to generate pseudo samples that capture task-specific distributions. It incorporates distillation to reduce noise in these samples\highlight{, outperforming baselines like Experience Replay and HAT in intent detection and slot filling.} PP \cite{razdaibiedina2023progressive} mitigates catastrophic forgetting \highlight{and enables forward transfer} without data replay or excessive task-specific parameters. This method appends new soft prompts to previously learned ones, keeping the base model unchanged. \highlight{It surpasses LPT5, IDBR, and MBPA++ on both standard and custom continual learning benchmarks.} Furthermore, ConTinTin \cite{yin2022contintin} provides a framework for sequentially mastering tasks through textual instructions, supporting both forward and backward transfer of knowledge. DYNAINST \cite{mok-etal-2023-large}, designed for lifelong in-context instruction learning, enhances a PLM's generalization by combining parameter regularization with experience replay. Its Dynamic Instruction Replay, using Dynamic Instance Selection (DIS) and Dynamic Task Selection (DTS), optimizes memory and computation by selectively replaying relevant instances and tasks.

\subsubsection{LLMs-based TIL}
\label{sect: LLMs-based TIL}
Recent attention has focused on the convergence of LLMs with CL methods, exemplified by significant contributions \cite{wang2023trace, peng2024scalable}. Benefiting from vast corpora and advanced hardware infrastructure, LLMs showcase remarkable capabilities in language comprehension and generation. However, challenges arise in scenarios involving sequential tasks, where LLMs often exhibit a decline in performance known as catastrophic forgetting.

\paragraph{Traditional Methods.}

DynaMind \cite{du2023static} (Figure \ref{fig:dynamind}) stands as a pioneering framework that enhances LLM output precision through the integration of memory mechanisms and modular operators. The framework consists of three key components: a memory module that stores and updates learned knowledge, a modular operator that processes incoming data, and a continual learning (CL) module that dynamically adjusts LLM parameters as new knowledge is introduced. In parallel, JARe \cite{peng2024scalable} incorporates Dynamic Task-related Knowledge Retrieval (DTKR) to enable adaptive adjustments of language models for specific downstream tasks. By leveraging task distributions within vector space, JARe aims to optimize the continual learning process, streamlining task-specific knowledge adaptation.

\paragraph{Parameter-Efficient Tuning Methods.}

Large language models (LLMs) face significant challenges that limit their practical applications, such as high computational and memory demands, and vulnerability to catastrophic forgetting. These issues highlight the need for more efficient and robust training and deployment strategies. ConPET \cite{song2023conpet}, designed for the continual adaptation of LLMs across diverse tasks, leverages parameter-efficient tuning (PET) strategies to enhance both efficiency and performance. ConPET operates in two modes: Static ConPET\highlight{, optimized for moderate-scale tasks,} and Dynamic ConPET\highlight{, tailored to large-scale tasks}. Additionally, ELM \cite{jang2023exploring} introduces a compact expert adapter trained for each task, with a retrieval mechanism that selects the most suitable expert LLM for new tasks. Furthermore, O-LoRA \cite{wang-etal-2023-orthogonal} \highlight{preserves task-specific knowledge by} using low-rank approximations of previous task gradients, while the orthogonality constraint \highlight{ensures task-specific subspaces remain distinct}, minimizing interference from new tasks. \highlight{This method demonstrates robustness across various task sequences.}

\paragraph{Instruction Tuning-based Methods.}

Continual-T0 \cite{scialom2022continual} incorporates rehearsal techniques alongside \highlight{instruction tuning of LLMs}, showing strong performance across 70 datasets. However, its generalization to underrepresented languages remains limited. In response, InstructAlign \cite{cahyawijaya2023instruct} proposes aligning new languages with those previously learned, leveraging abundant linguistic resources to mitigate catastrophic forgetting. The key innovation lies in enhancing language adaptation for instruction-tuned LLMs, particularly by incorporating underrepresented languages. \highlight{Unlike traditional methods focused on single-task learning, RoboCoder \cite{li2024robocoder} enables robots to tackle a wide range of increasingly complex tasks by continuously refining and updating action codes based on environmental feedback.} \highlight{EUREKA \cite{ma2024eureka} operates by leveraging LLMs to generate reward functions from the environment’s source code without requiring pre-defined templates or task-specific prompts. Through its evolutionary approach and in-context learning capabilities, it continually refines the rewards to achieve superior performance across diverse RL tasks.}

\subsubsection{VLMs-based TIL}
\label{sect: VLMs-based TIL}
The long-term sustainability of pre-trained visual-language models (VLMs) is increasingly under scrutiny due to their dependence on continually expanding datasets. Although these models demonstrate robust performance across a diverse range of downstream tasks, the incessant growth of real-world data poses substantial challenges to the sustainability of traditional offline training methodologies.

\paragraph{Traditional Methods.}
CTP \cite{zhu2023ctp} employs topology preservation and momentum contrast to maintain consistent relationships within sample mini-batches across tasks, thereby preserving the distribution of prior embeddings. ZSCL \cite{zheng2023preventing} addresses the challenge of zero-shot transfer degradation in VLMs by employing a label-free dataset for distillation in the feature space, alongside weight regularization in the parameter space. Additionally, ZSCL has been adapted for the CIL setting, expanding its applicability to a wider range of continual learning scenarios. \highlight{SND \cite{yu2024select} tackles catastrophic forgetting and zero-shot performance degradation in VLMs by leveraging a dual-teacher model. The pre-trained VLM retains zero-shot generalization, while the fine-tuned VLM preserves task-specific knowledge. A teacher selection mechanism quantifies discrepancies between their feature representations. Similarly, AwoForget \cite{zhengadapt} employs a graph-based multi-modal proximity distillation framework to mitigate forgetting in VLMs. It constructs a cross-modal graph to represent relationships between image-text pairs and adopts a dual-teacher distillation approach for maintaining both zero-shot and task-specific knowledge. This method also extends its applicability to CIL.}

\paragraph{Parameter-Efficient Methods.}
MoE-Adapters4CL \cite{yu2024boosting} introduces a parameter-efficient continual learning method to mitigate long-term forgetting in incremental learning of VLMs. Their approach involves dynamically extending the pre-trained CLIP model to accommodate new tasks by integrating a Mixture-of-Experts (MoE) adapter. Specifically, MoE consists of several LoRA adapter experts and routers, where the router calculates gating weights and uses the $TopK$ function to select the $k$ most relevant experts for learning the current task. To maintain the zero-shot recognition capabilities of the visual-language model, a Distribution Discriminative Automatic Selector (DDAS) is further introduced, which can automatically route in-distribution and out-of-distribution inputs to the MoE adapters and the original CLIP, respectively. Furthermore, the MoE-Adapters4CL framework has also been adapted for use in the CIL setting.

\paragraph{Instruction Tuning-based Methods.}
By decoupling prompts and prompt interaction strategies, TRIPLET \cite{qian2023decouple} (Figure \ref{fig:triplet}) effectively captures complex interactions between modalities. This includes specific designs for visual, textual, and fused prompts, as well as how to interact between different tasks through these prompts and retain crucial information, thereby reducing catastrophic forgetting. Decoupled prompts are designed to separate prompts in terms of multi-modality, layer-wise, and complementary, with each type of prompt containing learnable parameters intended to capture modality-specific knowledge from pre-trained VLMs and training data. 


\subsection{Class-Incremental Learning}
\label{sect: Class-Incremental Learning}
\subsubsection{PLMs-based CIL}
\label{sect: PLMs-base CIL}

\paragraph{Traditional Methods.}

\begin{figure}[t!]
\vspace{-2mm}
    \centering
    \begin{subfigure}[b]{0.3\textwidth}
        \centering
        \includegraphics[width=\textwidth]{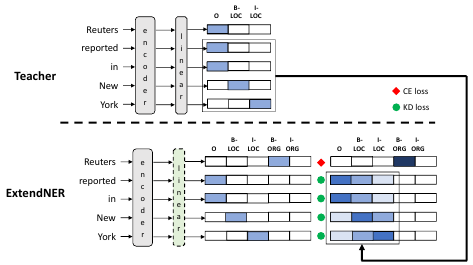}
        \caption{ExtendNER}
        \label{fig:extendner}
    \end{subfigure}
    \hfill
    \begin{subfigure}[b]{0.3\textwidth}
        \centering
        \includegraphics[width=\textwidth]{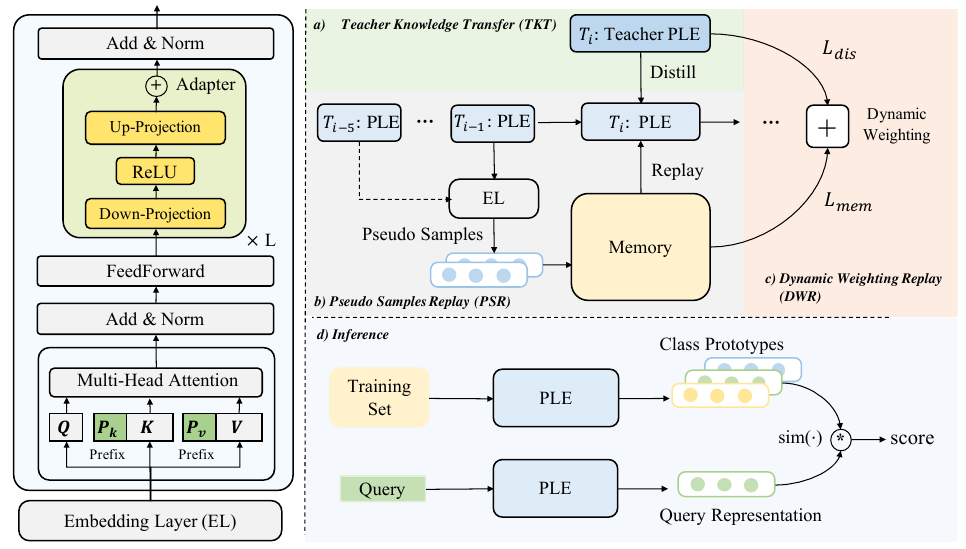}
        \caption{PLE}
        \label{fig:ple}
    \end{subfigure}
    \hfill
    \begin{subfigure}[b]{0.3\textwidth}
        \centering
        \includegraphics[width=\textwidth]{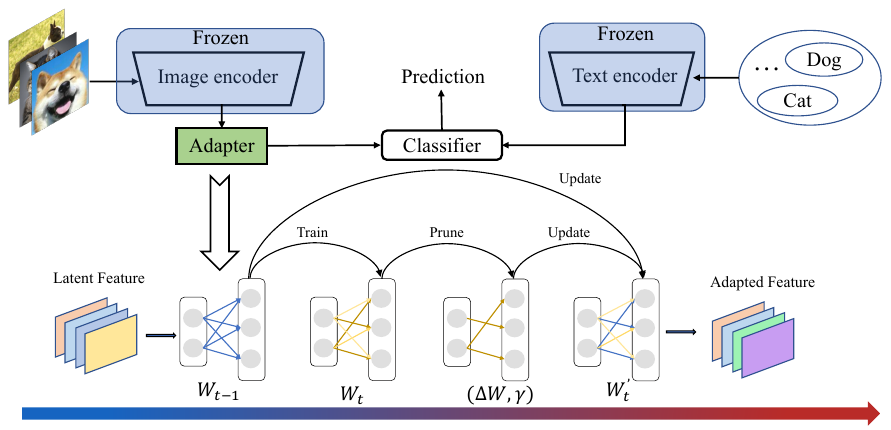}
        \caption{Adaptation-CLIP}
        \label{fig:adaptation-clip}
    \end{subfigure}
        \vspace{-3mm}
    \caption{Frameworks in CIL: ExtendNER (PLM-based) \cite{monaikul2021continual}, PLE (PLM-based) \cite{li2022continual}, Adaptation-CLIP (VLM-based) \cite{liu2023class}.}
    \label{fig:main3}
        \vspace{-4mm}
\end{figure}

ExtendNER \cite{monaikul2021continual} (Figure \ref{fig:extendner}) introduces a continual learning framework for Named Entity Recognition (NER) that reduces re-annotation efforts. It uses knowledge distillation, where a "teacher" NER model helps a "student" model learn to recognize new entity types while retaining knowledge of previous ones. SRC \cite{liu2019continual} focuses on sentence representation learning by initializing corpus-independent encoders and refining them through Boolean operations on conceptor matrices. IDBR \cite{huang2021continual} improves continual text classification by disentangling task-generic and task-specific representations. It uses regularization techniques and auxiliary tasks like next sentence prediction to enhance knowledge retention and generalization\highlight{, outperforming many existing methods across various task sequences.}

\paragraph{Instruction Tuning-based Methods.}
Instead of retaining examples, PAGeR \cite{varshney2022prompt} uses PLMs to generate intent-specific utterances for new tasks while maintaining performance on previous ones. It selectively preserves relevant contexts as prompts, ensuring efficient memory usage. \highlight{By utilizing PLMs, PAGeR is effective in real-world scenarios where new intents and relations are continuously introduced, offering a memory-efficient alternative to traditional methods.}

\paragraph{Parameter-Efficient Tuning Methods.}

In task-oriented dialogue systems, continuous few-shot intent detection (CFID) focuses on recognizing new intents with a few examples. To address this, PLE \cite{li2022continual} (Figure \ref{fig:ple}) employs a parameter-efficient tuning approach that integrates a Continual Adapter module with a frozen PLM and a Prefix-guided Attention mechanism. \highlight{This design effectively reduces forgetting while minimizing the number of trainable and stored parameters, making it an ideal choice for real-world applications where computational efficiency is critical.} Similarly, EPI \cite{wang2023rehearsal} allocates task-specific private parameters alongside a shared model, ensuring precise retrieval and superior performance in continual learning. It reduces storage needs through random static masking \highlight{ and consistently outperforms rehearsal-free methods, even competing with rehearsal-based ones on far-domain datasets.} For environments with sequentially presented data, DE\&E \cite{wojcik2023domain} uses a mixture of binary class-specific experts, dynamically selecting the best expert for each input, facilitating incremental learning by combining expert outputs for final classification.

\subsubsection{VLMs-based CIL}
\label{sect: VLMs-based CIL}

\paragraph{Traditional Methods.}

VLM-PL \cite{kim2024vlm} leverages a VLM to enhance the pseudo-labeling process, integrating new object classes into a detection model without compromising previously learned categories. GMM \cite{cao2024generative} leverages LLMs for class-incremental learning. This innovative approach entails the \highlight{generation of labels for images by employing an adapted generative model.}  PROOF \cite{zhou2023learning} maps pre-trained features into a new feature space designed to preserve prior knowledge. To enhance the use of cross-modal information, \highlight{it incorporates a fusion module with an attention mechanism, allowing the model to retain essential features when adapting to new tasks.} CLAP \cite{jha2024clap4clip} offers a new method for adapting VLMs to novel tasks without losing previously learned knowledge. This approach employs a Variational Inference framework to probabilistically model the distribution of visual-guided text features, enhancing fine-tuning reliability by accounting for uncertainties in visual-textual interactions.


\paragraph{Parameter-Efficient Tuning Methods.}

Adaptation-CLIP \cite{liu2023class} (Figure \ref{fig:adaptation-clip}) employs three strategies for continual learning: a linear adapter, a self-attention adapter, and prompt tuning. The adapters add layers to the image encoder while freezing the rest of the model, and prompt tuning enhances the text encoder by integrating new and prior prompts to maintain task comprehension. \highlight{DIKI \cite{tang2024mind} prevents the disruption of pre-trained knowledge by using a residual attention mechanism that injects new knowledge into a frozen backbone, minimizing interference. A distribution-aware calibration scheme is employed to maintain the zero-shot capabilities of VLMs.} \highlight{RAPF \cite{huang2024class} addresses class-incremental learning challenges using a pre-trained CLIP model with a linear adapter. It minimizes interference by identifying neighboring categories and applying hinge loss, while generating old class features from a Gaussian distribution to aid classification.} \highlight{STAR-Prompt \cite{menabue2024semantic} introduces a two-level prompting mechanism to balance stability and plasticity in large models like CLIP and ViT, reducing catastrophic forgetting while enabling continuous learning.}

\paragraph{Instruction Tuning-based Methods.}

LGCL \cite{khan2023introducing} introduces two key innovations: a refined prompt pool key query mechanism and category-level language guidance. \highlight{The former} leverages CLS features to optimize prompt selection with dynamic task-level mappings, improving accuracy and robustness\highlight{, while the latter} aligns output features with specific language representations in the vision transformer, enhancing task handling and model performance.



\section{Online Continual Learning}
\label{sect: Online Continual Learning}

\subsection{Hard Task Boundary}
\label{sect: Hard Task Boundary}

\subsubsection{PLMs-based HTB}
\label{sect: PLMs-based HTB}
The Hard Task Boundary (HTB) setting has been developed to enable continuous knowledge acquisition by learning models from a dynamically changing stream of textual data, without the need for dataset identifiers. 
For example, ProgModel \cite{shen2019progressive} have implemented HTB in slot filling, TAP-SLDA \cite{michieli2023online} have utilized it in audio classification, and AOS \cite{vander2023rehearsal} have explored its use in automatic speech recognition.

\begin{figure}[t!]
    \centering
    \begin{subfigure}[b]{0.45\textwidth}
        \centering
        \includegraphics[width=\textwidth]{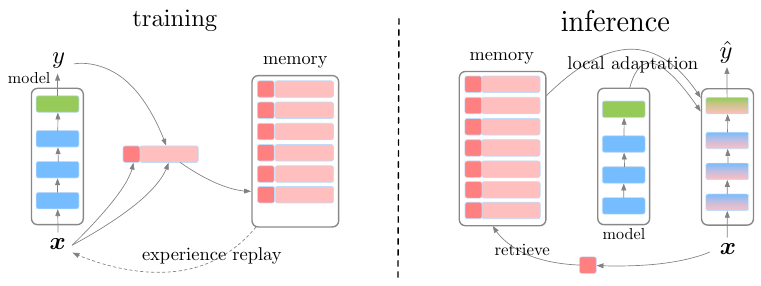}
        \caption{MBPA++}
        \label{fig:mbpa}
    \end{subfigure}
    \hspace{3em} 
    \begin{subfigure}[b]{0.32\textwidth}
        \centering
        \includegraphics[width=\textwidth]{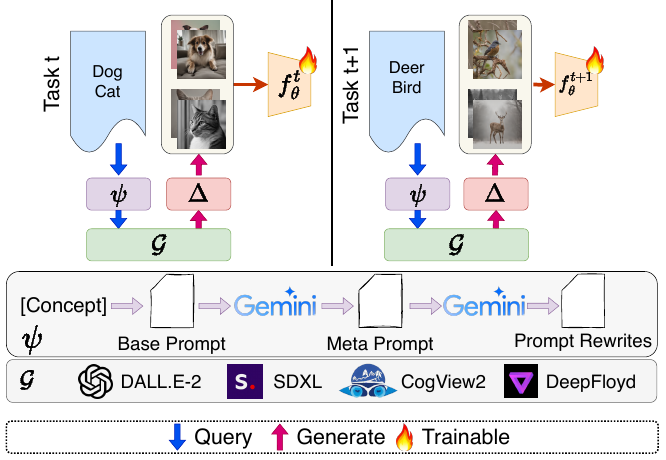}
        \caption{G-NoCL}
        \label{fig:G_NoCL}
    \end{subfigure}
    \vspace{-3mm}
    \caption{Frameworks in Online Continual Learning: MBPA++ (PLM-based HTB/BTB) \cite{de2019episodic}, G-NoCL (VLM-base BTB) \cite{seo2024just}.}
    \label{fig:main4}
    \vspace{-5mm}
\end{figure}

\paragraph{Traditional Methods.}
Continual learning (CL) methodologies, particularly pertinent to online scenarios, encompass a variety of approaches. These include parameter-isolation-based methods \cite{de2019episodic,wang2020efficient}, replay-based methods \cite{holla2020meta} and regularization-based methods \cite{vander2023rehearsal,liu2021lifelong1}.
MBPA++ \cite{de2019episodic} (Figure \ref{fig:mbpa}) introduces a framework for lifelong language learning, enabling a pre-trained model to learn continually from textual examples without requiring labeled datasets. It employs an episodic memory system with sparse experience replay and local adaptation techniques to prevent catastrophic forgetting. Extending this framework, Meta-MBPA++ \cite{wang2020efficient} integrates three core lifelong learning principles, enhancing performance in text classification and question-answering tasks while using only 1\% of the typical memory usage. CID \cite{liu2021lifelong1} is devised for lifelong intent detection. This method uses cosine normalization, hierarchical knowledge distillation, and inter-class margin loss to tackle the challenges of data imbalances in the lifelong intent detection task, aiming to mitigate the negative impacts associated with these imbalances.

\subsubsection{VLMs-based HTB}
\label{sect: VLMs-based HTB}

\highlight{PEGP \cite{qiao2024gradient} utilizes various parameter-efficient tuning methods and an orthogonal gradient projection mechanism to prevent catastrophic forgetting in continual learning. Using backbone networks like ViT and CLIP, PEGP ensures that gradient updates during training are orthogonal to the feature subspace of previous tasks, preserving old knowledge while learning new ones.}

\subsection{Blurry Task Boundary}
\label{sect: Blurry Task Boundary}

\subsubsection{PLMs-based BTB}
\label{sect: PLMs-based BTB}

MBPA++ \cite{de2019episodic} and Meta-MBPA++ \cite{wang2020efficient} exemplify models capable of adapting to environments with indistinct task boundaries. TPEM \cite{geng2021continual} adopts a tripartite approach within an encoder-decoder framework, utilizing pruning, expanding, and masking techniques. Pruning preserves essential information from previous tasks, expansion increases model capacity for new tasks, and masking mitigates interference from prior task weights, thereby enhancing learning efficiency. Online meta-learning (OML) \cite{javed2019meta} and the neuromodulatory meta-learning algorithm (ANML) \cite{beaulieu2020learning} are initially designed to learn sequences of tasks during the testing phase. Holla et al. \cite{holla2020meta} adapts these algorithms for conventional continual learning, focusing on performance evaluation on previous tasks. Their enhanced versions, OML-ER and ANML-ER, incorporate an episodic memory module for experience replay. \highlight{The S6 framework addresses the challenges of continual relation learning in real-world scenarios, where data arrives in a streaming manner with noisy labels and shifting distributions.}

\subsubsection{VLMs-based BTB}
\label{sect: VLMs-based BTB}
DKR \cite{cui2024continual} is devised to mitigate the propagation of incorrect information in foundation LMs. It operates by initially leveraging an existing model to identify and exclude obsolete or erroneous knowledge when confronted with new data. Subsequently, a rectification process is employed to amend these inaccuracies while preserving valid data associations. 
\highlight{SIT \cite{wang2024clip} addresses the challenge of asymmetry in Online Lifelong Learning (OLL) with VLMs, where the image encoder is tuned on the current batch of data, while the text encoder can access features from previously seen classes. SIT reformulates the loss function by restricting the text features in the comparison to only those present in the current batch, ensuring that both image and text feature updates remain balanced. }

\highlight{OLiVia-Nav \cite{narasimhan2024olivia} is an innovative architecture designed for social navigation in human-centered environments like hospitals and offices, integrating VLMs with an online lifelong learning framework. The system employs a novel distillation approach, Social Context Contrastive Language Image Pre-training (SC-CLIP), to transfer the social reasoning capabilities of large VLMs into lightweight models, enabling real-time encoding of social and environmental context. }
\highlight{Generative Name only Continual Learning (G-NoCL) \cite{seo2024just} (Figure \ref{fig:G_NoCL}) is a framework designed to overcome the challenges of manual annotation and web-scraped data in continual learning. The core of the model is the Prompt Refiner Module, which transforms basic concept names into diverse, context-rich prompts. }

\begin{table*}[t!]
\centering
\scriptsize
\caption{The statistics information of the existing CL datasets. \#D/T/C means the number of domains/tasks/classes for DIL, TIL and CIL, respectively.}
\label{tab:datasets}
\vspace{-3mm}
\setlength{\tabcolsep}{1.5pt} 
\begin{tabular}{p{1.8cm}cccccp{1.8cm}p{4cm}cc}
\hline
\textbf{Datasets}               & \textbf{\#Train} & \textbf{\#Val} & \textbf{\#Test} & \textbf{\#Total} & \textbf{CL Settings}        & \textbf{NLP Problems}  & \textbf{Source}                                                                                                                                                                           & \textbf{Language} & \textbf{\#D/T/C} \\ \hline
\multicolumn{9}{c}{Offline} \\ \hline

Progressive Prompts \cite{razdaibiedina2023progressive}            &       -           &       -         &             -    &             -     & TIL   & Mixed 15 classification tasks  &    AGNews, Amazon
Reviews, Yelp Reviews, DBpedia, Yahoo Answers, MNLI, QQP, RTE, SST2, WiC, CB, COPA, MultiRC, BoolQ                                                       & English           & 15 tasks      \\
MeLL \cite{wang2021mell} & 1,430,880           & 173,781          & 118,240           & 1,722,901           & TIL & Intent classification            & Snips, TOP semantic parsing, Facebook Multilingual Task Oriented Dataset, e-commerce dialogue system AliMe                                                                   & English           & 1184 tasks      \\ 
Continual-T0 \cite{scialom2022fine}                    & 800,000          & -              & 33,382          & 833,382          & TIL & Language generation task & WikiAuto, Gigaword,  Subreddit haiku, ELI5, Empathetic dialogues, e-SNLI, Twitter & English           & 8 tasks       \\
COPF \cite{zhang2023copf}                    & -          & -              & -          & -          & TIL & Value alignment & IMDB, HH-RLHF, Reddit TL, DR & English           & 3 tasks       \\
ACM \cite{zhang2022continual}                              & 50,725           & -          & 27,944           & 78,669           & TIL & Mixed 4 generation tasks & E2E NLG, RNNLG, WikiSQL, CNN/DailyMail, MultiWOZ  & English           & 4 tasks      \\
CODETASKCL \cite{yadav2023exploring}                    & 181,000          & 9,700              & 10,000          & 200,700          & TIL &  Mixed 4 code-centric tasks & CONCODE, CodeTrans, CodeSearchNet, BFP & Hybrid          & 4 tasks       \\
Lifelong SimpleQuestions \cite{wang-etal-2019-sentence}                    & -          & -              & -          & -          & TIL &  Relation extraction & SimpleQuestions & English          & 20 tasks       \\ 
Lifelong FewRel \cite{wang-etal-2019-sentence}                    & -          & -              & -          & -          & TIL & Relation extraction & FewRel & English          & 10 tasks       \\ 
InstrDialog \cite{zhang2023citb}                              & 9,500           & 950          & 1,900           & 12,350           & TIL & Mixed 3 dialogue tasks  & SuperNI                                                                               & English           & 19 tasks      \\
InstrDialog++ \cite{zhang2023citb}                              & 3,800           & 1,900          & 3,800           & 9,500           & TIL & Mixed 22 dialogue tasks  & SuperNI                                                                             & English           & 38 tasks      \\
ConTinTin \cite{yin2022contintin}                    & -          & -              & -          & -          & TIL & Mixed classification and generation tasks & NATURAL-INSTRUCTIONS & English           & 61 tasks       \\
Conure \cite{10.1145/3404835.3462884}                    & -          & -              & -          & -          & TIL &  User
representations & Tencent TL, Movielens &  English           & 9 tasks \\
NAVER Shopping \cite{kim2023task}                   & -           & -           & -          & -           & TIL                            &  Search query prediction & Tencent TL, Movielens & English           & 6 tasks               \\

TRACE \cite{wang2023trace}                          & 40,000           & -              & 16,000          & 56,000           & TIL   & Mixed diverse tasks   & ScienceQA, FOMC, MeetingBank, C-STANCE, 20Minuten, CodeXGLUE, NumGLUE    & Hybrid            & 8 tasks       \\
ABSC \cite{ke2021adapting} & 3,452            & 150            & 1,120           & 4,722            & DIL & Aspect-based sentiment classification & L5Domains, Liu3Domains, Ding9Domains, SemEval14    & English           & 19 domains      \\
LAMOL \cite{sun2019lamol}                              & 284,824           & -          & 39,716           & 324,540           & DIL & Mixed classification and labeling tasks  & AGNews, Amazon Reviews, Yelp Reviews, DBpedia, Yahoo Answers, SQuAD, WikiSQL, SST, QA-SRL, WOZ                                                                               & English           & 5 domains      \\ 
RVAE\_LAMOL \cite{wang2022rvae}                   & 15,870           & -           & 5,668          & 21,538           & DIL                            &    Generation tasks & WOZ, QA-SRL, SST & English           & 3 domains               \\
COPF \cite{zhang2023copf}                    & -          & -              & -          & -          & DIL & Value alignment & Standard Human Preference (SHP) & English           & 18 domains       \\
C-PT \cite{zhu2022continual}                              & 38,745           & 5,210          & 11,349           & 40,287           & DIL & Dialogue state tracking    & Schema-Guided Dialog dataset (SGD)  & English           & 19 domains      \\ 
CPT \cite{ke2022continualb}                             & 3,121,926           & -          & -           & 3,121,926           & DIL & Domain-adaptive pre-training task    & Yelp Restaurant, AI Papers, ACL Papers, AGNews, SemEval-res, ACL-ARC, SCIERC                                                                              & English           & 4 domains      \\
CKL\cite{jang2021towards}                             & -           & -          & -           & 30,372           & DIL & Domain-adaptive pre-training task   & INVARIANTLAMA, UPDATEDLAMA, NEWLAMA, NEWLAMA-EASY & English           & 3 domains      \\
ELLE \cite{qin2022elle}                             & -           & -          & -           & -           & DIL & Domain-adaptive pre-training task & BOOKCORPUS(WB), NEWS ARTICLES(NS), AMAZON REVIEWS(REV), BIOMEDICAL PAPERS(BIO), COMPUTER SCIENCE PAPERS(CS)   & English           & 5 domains      \\
Domain-incremental Paper Stream \cite{jin2021lifelong} & - & - & - & - & DIL & Mixed 2 information extraction tasks & S2ORC & English & 4 domains \\
Chronologically-ordered Tweet Stream \cite{jin2021lifelong} & - & - & - & - & DIL & Mixed 2 classification tasks & Tweets & English & 4 domains \\
AdapterCL \cite{madotto2020continual}                              & 31,426           & 4,043          & 4,818           & 40,287           & DIL & Mixed 4 dialogue tasks &  TaskMaster2019(TM19), TaskMaster2020(TM20), Schema Guided Dialogue(SGD), MultiWoZ   & English           & 37 domains      \\
DE\&E \cite{wojcik2023domain}                   & 28,982           & -           & 12,089          & 41,071           & CIL                            &    Text classification & BBC News, Newsgroups, Complaints & English           & 3 tasks   \\

EPI \cite{wang2023rehearsal}                              & 12,840           & 3,524          & 6,917           & 23,281           & CIL & Mixed 2 classification tasks & AGNews, Amazon Reviews, Yelp Reviews, DBpedia, Yahoo Answers, Web of Science (WOS) & English           & 13 classes      \\
PAGeR \cite{varshney2022prompt}                              & 59,754           & 7,115          & 15,304           & 82,173           & CIL & Intent classification & CLINC150, HWU64, BANKING77, Stackoverflow S20, SGD, MWOZ, FewRel & English           & 355 classes      \\
PLE \cite{li2022continual}                              & 4,669           & 4,650          & 31,642           & 40,961           & CIL & Intent classification & CLINC150, ATIS, HWU64, BANKING77, MTOP, SNIPS, LEYZER, MSLU, TOP & English           & 477 classes      \\ 
ExtendNER \cite{monaikul2021continual}                              & 130,495           & 22,717          & 16,077           & 169,289           & CIL & Named Entity Recognition                                                                             & CoNLL-03, OntoNotes & English           & 4 classes      \\
\hline
\multicolumn{9}{c}{Online} \\ \hline
MBPA++ \cite{de2019episodic} & 115,000           & -          & 7,600           & 122,600           & Hard and Blurry &    Mixed 5 classification tasks & AGNews, Amazon Reviews, Yelp Reviews, DBpedia, Yahoo Answers & English           & 5 tasks      \\ 
MBPA++ \cite{de2019episodic}                   & 306,000           & 35,000           & -          & 341,000           & Hard and Blurry                            &    Question answering & QuAC, SQuAD, Trivia Wikipedia, Trivia Web & English           & 4 tasks               \\
Lifelong FewRel \cite{holla2020meta}                    & -          & -              & -          & -          & Hard and Blurry & Few-shot relation detection & FewRel & English          & 10 tasks \\ 
Firehose \cite{hu2022drinking}                   & -           & -           & -          & 110,000,000           & Blurry                            &Personalized online language learning & Twitter & English           & 1 tasks          \\
TemporalWiki \cite{jang2022temporalwiki}                   & -           & -           & -          & -           & -     & - & Wikipedia
 & English           & -               \\
\hline

\end{tabular}
\end{table*}

\section{Datasets}
\label{sect: Datasets}
In this section, \highlight{we review the typical datasets for offline and online continual learning for different tasks} (Table \ref{tab:datasets}). 

\subsection{Offline Datasets for NLP}
\subsubsection{Datasets for Classification.}

\paragraph{Text Classification.} The most typical task for continual learning is text classification. 
The foundational text classification benchmark encompasses five text classification datasets introduced by \cite{zhang2015character}, including AG News, Amazon Reviews, Yelp Reviews, DBpedia, and Yahoo Answers \cite{sun2019lamol}. Particularly, the AG News dataset has 4 classes for news classification; the Amazon and Yelp dataset has 5 classes for sentiment analysis; the DBpedia dataset has 14 classes for Wikipedia text classification; \highlight{and the Yahoo dataset has 10 classes for Q\&A classification. 
Building upon this,} Razdaibiedina et al. \cite{razdaibiedina2023progressive} developed a novel continual learning (CL) benchmark. 
This benchmark not only utilizes the foundational text classification benchmark but also integrates additional datasets from \highlight{the GLUE benchmark \cite{wang2018glue}, SuperGLUE benchmark \cite{wang2019superglue}, and the IMDB dataset \cite{maas2011learning}. 
DE\&E \cite{wojcik2023domain} uses three common text classification data sets} with different characteristics-News-groups, BBC News, and Consumer Finance Complaints2. Such datasets can be used to evaluate the models on tasks with different difficulty levels.

The datasets introduced in \cite{wang2023rehearsal} further are categorized into two groups based on the domain relevance between tasks: far-domain and near-domain. The far-domain group comprises two text classification tasks, which are foundational benchmarks \cite{zhang2015character} divided into topic classification (AG News, Yahoo Answers, DBpedia) and sentiment classification (Yelp, Amazon Reviews). In contrast, the near-domain group uses the Web of Science (WOS) \cite{kowsari2017hdltex} and 20 Newsgroups \cite{lang1995newsweeder}, \highlight{which are restructured according to their high inter-task relevance. }

\paragraph{Intent Classification.} 
Some studies focus on intent classification tasks, where the classes are quite different in \highlight{various domains or scenarios.
The dataset,} as introduced in PAGeR \cite{varshney2022prompt}, aims to tackle the lifelong intent detection problem by combining three public intent classification datasets (CLINC150 \cite{larson2019evaluation}, HWU64 \cite{liu2021benchmarking}, BANKING77 \cite{casanueva2020efficient}), one text classification dataset (Stackoverflow S20 \cite{xu2017self}), and two public multidomain dialog intent detection datasets (SGD \cite{rastogi2020towards}, MWOZ \cite{budzianowski2018multiwoz}). Moreover, FewRel \cite{han2018fewrel} is also incorporated to tackle the lifelong relation extraction problem. This integration is intended to simulate real-world applications by encompassing a broad spectrum of domains and query distributions, thereby facilitating the development of more robust and versatile intent detection systems. 

Conversely, the dataset compiled in PLE \cite{li2022continual} consolidates nine well-regarded intent detection datasets, including CLINC150 \cite{larson2019evaluation} and HWU64 \cite{liu2021benchmarking}, among others, \highlight{arranged in a fixed random sequence to form a standardized benchmark. 
The dataset described by MeLL} \cite{wang2021mell} specifically addresses intent detection within two distinct contexts: task-oriented dialogues (TaskDialog-EUIC) and real-world e-commerce interactions (Hotline-EUIC). TaskDialog-EUIC integrates data from Snips \cite{coucke2018snips}, TOP semantic parsing \cite{gupta2018semantic}, and Facebook's Multilingual Task Oriented Dataset \cite{schuster2018cross} into 90 tasks with overlapping label sets, \highlight{amounting to over ten thousand samples. }

\paragraph{Fine-grained Sentiment Analysis.} 
Ke et al. \cite{ke2021adapting} develop a task incremental learning dataset for aspect-based sentiment classification (ABSC). This dataset aggregates reviews from four distinct sources, thereby enhancing its diversity and applicability across multiple domains. 
\highlight{The sources include the L5Domains dataset by Hu et al. \cite{hu2004mining}, the Liu3Domains dataset \cite{liu2015automated}, the Ding9Domains dataset \cite{ding2008holistic}, and the SemEval14 dataset.}

\subsubsection{Datasets for Generation.}
Diverse datasets function as crucial benchmarks for exploring various dimensions of language and \highlight{code generation. 
A particularly} significant dataset highlighted in the work by Continual-T0 \cite{scialom2022fine} focuses on English language generation tasks, including text simplification and empathetic dialogue generation, among others \cite{camburu2018snli,DVN/JBXKFD_2017}. 
In a subsequent study, Luo et al. \cite{luo2023empirical} conduct an analysis of catastrophic forgetting on Bloomz \cite{scao2022bloom} using Continual T0 datasets.
The dataset, introduced in LAMOL \cite{sun2019lamol}, integrates elements from both DecaNLP \cite{mccann1806natural} and the foundational text classification \highlight{benchmark \cite{zhang2015character}. 
Moreover, the dataset} devised in RVAE\_LAMOL \cite{wang2022rvae}, employs three tasks from DecaNLP: the English Wizard of Oz (WOZ) for goal-oriented dialogue, QA-SRL for semantic role labeling in a SQuAD-style format, and SST, \highlight{which is a binary version of the Stanford Sentiment Treebank.} 

The dataset introduced in COPF \cite{zhang2023copf} represents a pioneering effort in \highlight{applying both TIL and DIL within} the context of benchmarks that utilize existing human preferences. 
Specifically, the TIL framework in this dataset mandates that the model sequentially acquires knowledge from three distinct tasks. These include the question-answering task utilizing the HH-RLHF dataset \cite{bai2022training}, the summarization task based on the Reddit TL, DR dataset with human feedback \cite{volske-etal-2017-tl}, and the positive film review generation task using the IMDB dataset \cite{maas2011learning}. Meanwhile, the DIL framework requires the model to adapt to three distinct segments from the SHP dataset, as described by Ethayarajh et al. \cite{ethayarajh2022understanding}. 

The dataset described in ACM \cite{zhang2022continual} explores sequence generation and categorizes tasks into ``similar" and ``dissimilar" groups based on their characteristics. Tasks classified as similar, including E2ENLG \cite{novikova2017e2e} and four domains (restaurant, hotel, TV, laptop) from RNNLG \cite{wen2015semantically}, demonstrate shared patterns and are tested across four sequence orders, comprising a total of five tasks. In contrast, dissimilar tasks \highlight{such as WikiSQL \cite{zhong2017seq2sql}, CNN/DailyMail \cite{see2017get}, and MultiWOZ \cite{budzianowski2018multiwoz} exhibit significant} distributional shifts from previously encountered tasks. 
The CODETASKCL dataset, explored by Yadav et al. \cite{yadav2023exploring}, encompasses a diverse array of code-centric tasks, including code generation \cite{iyer2018mapping}, summarization \cite{husain2019codesearchnet}, translation \cite{lu2021codexglue}, and refinement \cite{tufano2019empirical} across various programming languages. This dataset significantly enhances the breadth of language processing applications within technical fields. 

\subsubsection{Datasets for Information Extraction.}
The dataset introduced in ExtendNER \cite{monaikul2021continual}, exemplifies a continual learning approach to Named Entity Recognition (NER). This dataset amalgamates the CoNLL-03 English NER \cite{sang2003introduction} and OntoNotes \cite{hovy-etal-2006-ontonotes}, covering a broad spectrum of entity types and sources. 
This hybrid dataset is structured to challenge the adaptability and generalization capabilities of NER systems across varied contexts.
Unlike the static nature of text in NER tasks, the Schema-Guided Dialog (SGD) \cite{rastogi2020towards} dataset, utilized in C-PT \cite{zhu2022continual}, serves the Dialog State Tracking aspect of IE, which involves maintaining the context of a dialog over time. The SGD dataset features 44 services across 19 domains, each treated as a separate task, and is designed to evaluate models on their ability to manage and extract information across conversational turns.
Lastly, the lifelong SimpleQuestions and lifelong FewRel datasets, devised in  \cite{wang-etal-2019-sentence} is crafted for the task of relation extraction. It merges elements from the SimpleQuestions \cite{bordes2015large} and FewRel \cite{han2018fewrel} to form a lifelong learning benchmark that confronts the challenges of relation detection in a few-shot context. 

\subsubsection{Datasets for Continual Pre-training.}
In the realm of continual pre-training for LLMs, the development and utilization of specialized benchmarks play a pivotal role in evaluating and enhancing the effectiveness of continual learning systems. The dataset, introduced in CPT \cite{ke2022continualb}, primarily focuses on the continual post-training of LMs across a series of domain-specific, unlabeled datasets. It provides a rigorous test environment by using diverse corpora such as Yelp Restaurant Reviews \cite{xu2019bert}, AI and ACL Papers \cite{lo-etal-2020-s2orc}, and AGNews articles \cite{zhang2015character}. Its main objective is to gauge how well an LM can incrementally integrate domain-specific knowledge without forgetting previously learned information, thereby enhancing its few-shot learning capabilities in these domains.
Contrary to the datasets employed in CPT \cite{ke2022continualb}, which evaluate domain-specific adaptability and incremental learning, the CKL benchmark \cite{jang2021towards} is meticulously designed to measure the LM's ability to retain timeless knowledge, update obsolete information, and acquire new knowledge. It comprises subsets like INVARIANTLAMA, UPDATEDLAMA, and NEWLAMA, which are crafted to probe specific types of knowledge that an LM may encounter in its learning trajectory. 

Whereas the aforementioned two datasets assess more controlled dimensions of knowledge integration and retention, the dataset introduced in ELLE \cite{qin2022elle} focuses on the dynamic scenario of accumulating streaming data from diverse sources in a lifelong learning context. This dataset mirrors the real-world challenge of a LM that must continuously adapt to new data inflows from multiple domains, including BOOKCORPUS \cite{zhu2015aligning}, NEWS ARTICLES \cite{zellers2019defending}, AMAZON REVIEWS \cite{he2016ups}, BIOMEDICAL PAPERS \cite{lo-etal-2020-s2orc} and COMPUTER SCIENCE PAPERS \cite{lo-etal-2020-s2orc}. The benchmark evaluates the LM's capacity to effectively integrate new information from these varied sources over time, highlighting the essential need for LMs to evolve in response to continual data growth and shifts in data distribution.
Jin et al. \cite{jin2021lifelong} construct data streams to represent two prevalent types of domain shifts observed in practical scenarios. The first, a Domain-incremental Paper Stream, simulates the sequential evolution of research areas within academic papers, encompassing diverse disciplines such as biomedical and computer science. The second, a Chronologically-ordered Tweet Stream, models the temporal progression of tweets over time.

\subsubsection{Datasets for Hybrid Tasks.}
An increasing number of datasets are adopting a hybrid task approach that integrates multiple learning paradigms and task types, aimed at enhancing the adaptability of models. \highlight{AdapterCL \cite{madotto2020continual} introduces a dataset for task-oriented dialogue systems, which incorporates} four task-oriented datasets: TaskMaster 2019 \cite{byrne2019taskmaster}, TaskMaster 2020 \cite{byrne2019taskmaster}, Schema Guided Dialogue \cite{rastogi2020towards}, and MultiWoZ \cite{budzianowski2018multiwoz}. 
Continual Instruction Tuning Benchmark (CITB) \cite{zhang2023citb} extends the concept of continual learning by focusing on instruction-based NLP tasks. Built on the comprehensive SuperNI \cite{wang2022super} dataset, it includes over 1,600 tasks across diverse NLP categories. 
The ConTinTin \cite{yin2022contintin} is an adaptation of the NATURAL-INSTRUCTIONS dataset, specifically restructured to facilitate a CL framework. 

The dataset, used in Conure \cite{10.1145/3404835.3462884}, consists of Tencent TL (TTL) \cite{yuan2020parameter} and Movielens (ML). The TTL dataset is designed to address three item recommendation tasks and three user profiling tasks, whereas the ML dataset exclusively focuses on three item recommendation tasks. 
Furthermore, Kim et al. \cite{kim2023task} introduced the proprietary NAVER Shopping dataset, which builds upon the previously mentioned datasets. The NAVER Shopping dataset features six tasks: two for search query prediction, two for purchased item category prediction, and two for user profiling, all designed to meet real-world industry requirements.
Finally, the TRACE dataset, introduced by Wang et al. \cite{wang2023trace}, is specifically designed to bridge the existing gap in the evaluation of LLMs \highlight{within the CL framework, encompassing a wide range of complex and specialized tasks.} 

\subsection{Online Datasets for NLP}
\subsubsection{Datasets for Classification.}
The foundational text classification benchmark, as introduced by Zhang et al. \cite{zhang2015character} has traditionally been applied in offline CL settings. Recent advancements have adapted this benchmark for online CL, notably in studies such as MBPA++ \cite{de2019episodic} and OML-ER \cite{holla2020meta}

\subsubsection{Datasets for Generation.}
The dataset, used in MBPA++ \cite{de2019episodic}, comprises three distinct question-answering collections: SQuAD 1.1 \cite{rajpurkar-etal-2016-squad}, TriviaQA \cite{joshi-etal-2017-triviaqa}, and QuAC \cite{choi-etal-2018-quac}. SQuAD 1.1 is a reading comprehension dataset based on Wikipedia articles, designed to assess the ability to derive answers from structured text. TriviaQA consists of question-answer pairs developed by trivia enthusiasts, accompanied by corroborative evidence sourced from both the web and Wikipedia, testing the model’s capability to handle diverse information sources. QuAC adopts a dialog-style format in which a student queries about information in a Wikipedia article and a teacher responds using text directly from the article, challenging the model's interactive response generation. 

\subsubsection{Datasets for Information Extraction.}

The lifelong relation extraction benchmark, used in OML-ER \cite{holla2020meta}, is structured by Wang et al. \cite{wang-etal-2019-sentence} based on FewRel. Unlike the original application by Wang et al., the benchmark in OML-ER is adapted for online continuous learning scenarios.

\subsubsection{Datasets for Other Tasks.}

Hu et al. \cite{hu2022drinking} compile the Firehose dataset, consisting of 110 million tweets from over 920,000 users between January 2013 and September 2019. This dataset is split into FIREHOSE 10M and FIREHOSE 100M. 
TemporalWiki \cite{jang2022temporalwiki} addresses temporal misalignment by serving as a lifelong benchmark that trains and evaluates LMs using consecutive snapshots of Wikipedia and Wikidata. This methodology assists in assessing an LM's capacity to both retain previously acquired knowledge and assimilate new information over time.

\subsection{Offline CL Datasets for Multi-modal Tasks}
The P9D dataset \cite{zhu2023ctp} consists of over one million image-text pairs from e-commerce data, organized into nine industry sector-based training tasks. It includes 1,014,599 training pairs, 2,846 for cross-modal retrieval tests, and 4,615 query pairs with 46,855 gallery pairs for multi-modal retrieval. 
Qian et al. \cite{qian2023decouple} introduce two novel benchmarks for continual learning, namely CL-TDIUC and CL-VQA2.0, which are derived from the TDIUC \cite{kafle2017analysis} and VQA2.0 \cite{goyal2017making}, respectively. These benchmarks are categorized into three scenarios: the Continual Vision Scenario, which deals with new visual scenes; the Continual Language Scenario, focusing on new questions in existing scenes; and the Continual Vision-Language Scenario, addressing changes in both questions and visuals. 
DKR \cite{cui2024continual} comprises five benchmark datasets: MS-COCO Caption (MS-COCO) \cite{lin2014microsoft}, Flickr30K \cite{young2014image}, IAPR TC-12 \cite{grubinger2006iapr}, ECommerce-T2I (EC) \cite{yang2021m6}, and RSICD \cite{lu2017exploring}. Furthermore, two experimental scenarios are established. The first scenario involves a sequential processing of the datasets, specifically MS-COCO, Flickr30K, IAPR TC-12, EC, and RSICD, in that order. The second scenario, which builds on the approach proposed by Ni et al. \cite{ni2023continual}, partitions the EC dataset into five sub-datasets for the training phase. The model's performance is subsequently tested on the Flickr30K, MS-COCO, and EC datasets.


\highlight{LAION-400M \cite{schuhmann2021laion} and LAION-5B \cite{schuhmann2022laion} are both CLIP-filtered, large-scale, open-access datasets designed to support multimodal models like CLIP, but they differ significantly in scale and scope. LAION-400M contains 400 million image-text pairs primarily in English, while LAION-5B is a much larger and more diverse dataset with 5.85 billion pairs, including multilingual and language-agnostic examples.}
\highlight{CC3M \cite{sharma2018conceptual} and CC12M \cite{changpinyo2021conceptual} are large-scale datasets for vision-and-language pre-training, differing in scale, filtering, and diversity. CC3M has 3.3 million image-text pairs with strict filtering for high-precision captions, while CC12M, with 12.4 million pairs, relaxes these filters to capture a broader range of concepts, especially long-tail categories. }
\highlight{
COYO-700M \cite{byeon2022coyo} contains 747 million image-text pairs, designed to train various models with additional meta-attributes. }
\highlight{Datacomp \cite{gadre2024datacomp} is a benchmark designed to advance the creation of multimodal datasets, particularly for image-text models such as CLIP. 
}


\begin{figure*}
\vspace{-2mm}
    \centering
    \includegraphics[width=0.6\linewidth]{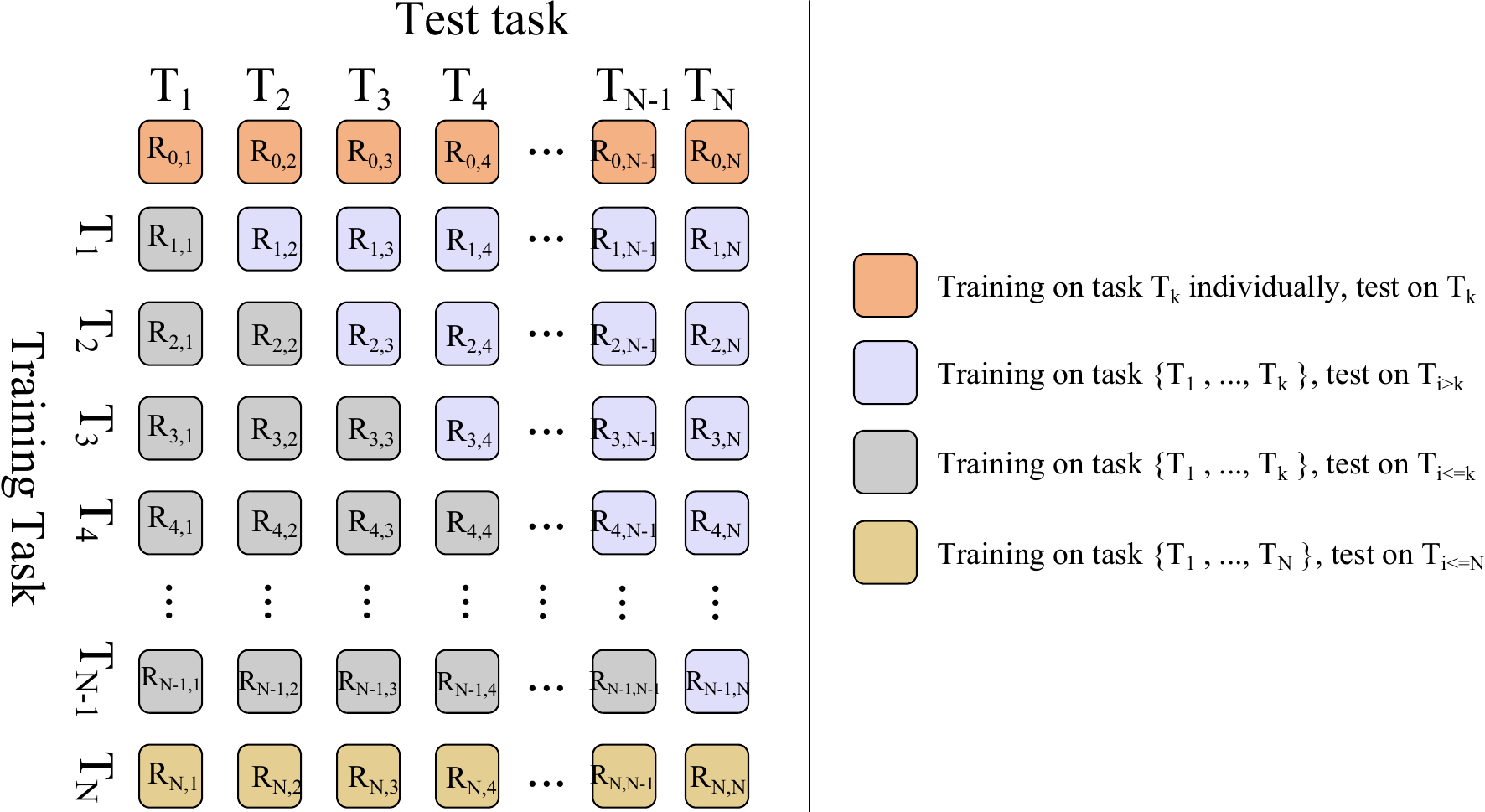}
   \vspace{-2mm}
    \caption{Illustration of calculating metrics.}
    \label{fig:evaluation}
    \vspace{-5mm}
\end{figure*}

\section{Metrics}
\label{sect: Metrics}
In this section, we review the principal metrics commonly used to evaluate continual learning. These metrics can be categorized into three main types: (1) overall performance, which assesses the algorithm's effectiveness across all tasks; (2) memory stability, which measures the extent to which an algorithm retains previously acquired knowledge; and (3) learning plasticity, which evaluates the algorithm's capacity to acquire new skills or knowledge. Each of these metrics provides insights into different aspects of the algorithm's performance in a continual learning context.

To begin, we establish the notation (Figure \ref{fig:evaluation}) used throughout the learning and evaluation phases of the model. Once the model completes a learning task, denoted as \(T_i\), it evaluates its performance on a test set that encompasses all \(N\) tasks, where \(N\) is the total number of tasks in the set \(T\). This evaluation is represented by a matrix \(R \in \mathbb{R}^{N \times N}\), wherein each element \(R_{i,j}\) indicates the model's test classification accuracy on task \(T_j\) after training on task \(T_i\).

\subsection{Overall Performance.} 
The metric termed ``Last" \cite{lopez2017gradient, zheng2023preventing} evaluates the overall performance of a continual learning (CL) method upon the completion of all tasks. Specifically, it computes the average score from the last row in the performance matrix \( R \).
\begin{equation}
    Last = \frac{1}{N} \sum_{i=1}^{N} R_{N,i}
\end{equation}

Also, Zheng et al. \cite{zheng2023preventing} devise the ``Avg" score metric, which computes the mean accuracy across all datasets and timestamps.
\begin{equation}
    Avg = \frac{1}{N} \sum_{i=1}^{N} \left(\frac{1}{N} \sum_{j=1}^{N} R_{i,j} \right)
\end{equation}

In the seminal works of Rebuffi et al. \cite{Rebuffi_2017_CVPR} and Douillard et al. \cite{douillard2020podnet}, the concept of Average Incremental Accuracy (AIA) is introduced. This metric is specifically designed to quantify the historical performance across different tasks. It calculates the average performance for each task by considering the lower triangular portion of the matrix $R$, effectively capturing the evolving competence of the system as new tasks are learned.
\begin{equation}
    AIA = \frac{1}{N} \sum_{i=1}^{N} \left(\frac{1}{i} \sum_{j=1}^{i} R_{i,j}\right)
\end{equation}

The metric, termed Transfer, is derived by computing the average of the performance values for tasks that are represented in the upper-right triangle of matrix $R$. This approach uniformly weights each dataset by averaging their performance across different tasks, thereby assessing the preservation of zero-shot transfer capabilities. Prior to commencing learning on task \( T_i \), no fine-tuning is performed on tasks that precede \( T_i \).
\begin{equation}
    Transfer = \frac{1}{N-1} \sum_{i=2}^{N} \left(\frac{1}{i-1} \sum_{j=1}^{i-1} R_{j,i}\right)
\end{equation}

Moreover, Chaudhry et al. \cite{chaudhry2018efficient} devise a metric known as Learning Curve Area (LCA), which quantifies the speed of learning in a model. 
Qin et al. \cite{qin2022elle} propose two metrics designed to evaluate pre-trained language models (PLMs) based on their performance within learned domains: Average Perplexity ($AP$) and Average Increased Perplexity ($AP^{+}$).

\subsection{Memory Stability.} 
Memory stability is typically evaluated \highlight{by backward transfer \cite{lopez2017gradient} and forgetting measure \cite{chaudhry2018riemannian}. Backward Transfer (BWT) emerges} as a pivotal concept extensively documented in the literature, notably by Lopez et al. \cite{lopez2017gradient} and Wu et al. \cite{wu2023online}. BWT measures the performance degradation on previously mastered tasks after the model is trained on new tasks. This performance degradation phenomenon is often referred to as ``forgetting".
\begin{equation}
    BWT = \frac{1}{N - 1} \sum_{i=1}^{N-1} R_{N,i} - R_{i,i}
\end{equation}

Additionally, Chaudhry et al. \cite{chaudhry2018riemannian} introduce the Forgetting Measure (FM), a metric designed to quantify the extent of forgetting a model experiences for a specific task. A lower FM indicates better retention of previous tasks. 
Davari et al. \cite{davari2022probing} \highlight{propose linear probes (LP) to assess representation forgetting via an optimal linear classifier trained on the frozen activations of a base network}. 
Representation forgetting is quantified by evaluating the change in Language Processing (LP) performance before and after the introduction of a new task. Kemker et al. \cite{kemker2018measuring} introduce three metrics, where \( \Omega_{\text{base}} \) assesses retention of initial learning, \( \Omega_{\text{new}} \) measures recall of new tasks, and \( \Omega_{\text{all}} \) evaluates overall proficiency in maintaining old knowledge and acquiring new information.
Additionally, researchers \cite{koh2022online} devise a novel metric, termed the Knowledge Loss Ratio (KLR), quantifies knowledge degradation using principles from information theory.

\subsection{Learning Plasticity.}
Evaluating learning plasticity can be effectively accomplished through two key metrics: forward transfer (FWT) \cite{lopez2017gradient} and intransigence measure (IM) \cite{chaudhry2018riemannian}.
Forward Transfer (FWT) \cite{lopez2017gradient} assesses the beneficial effects on the performance of subsequent tasks following a model's training on prior tasks.
\begin{equation}
    FWT = \frac{1}{N - 1} \sum_{i=2}^{N} R_{i-1,i} - R_{0, i}
\end{equation}
where \( R_{0, i} \) denotes the performance metric associated with training on task \( i \) independently. \highlight{Higher values of FWT indicate superior model performance.} 

Intransigence measure (IM), as defined by Chaudhry et al. \cite{chaudhry2018riemannian}, \highlight{quantifies a model's inability to learn new tasks by comparing the performance difference of a task when trained jointly with other tasks versus.} 
Moreover, Koh et al. \cite{koh2022online} introduce novel metrics, known as Knowledge Gain Ratio (KGR), \highlight{which quantifies the capacity to acquire new knowledge by calculating knowledge gain.}

\begin{table*}[t!]
\centering
\scriptsize
\caption{Comparison of typical continual learning methods, where Reg. and Para. mean Regularization and Parameter-isolation.}
\label{table: comparison of CL}
\vspace{-3mm}
\setlength{\tabcolsep}{0.5pt}{ 
\begin{tabular}{c|cccccc}
\hline
\textbf{CL settings} & \textbf{Method} & \textbf{FM Arch.} & \multicolumn{4}{c}{\textbf{Continual Learning Tech.}} \\ \cline{4-7} 
                  &                 &               & \textbf{Traditional} & \textbf{Continual Pre-training} & \textbf{Parameter-Efficient} & \textbf{Instruction Tuning} \\ \hline
\multirow{26}{*}{DIL}     & LFPT5 \cite{qin2022lfpt}  & T5 & \xmark & \xmark  & Replay | Reg. | Para. & \xmark           \\ 
                         & B-CL \cite{ke2021adapting}      & BERT       & \xmark & \xmark & Para. & \xmark             \\ 
                         & ELLE \cite{qin2022elle} & BERT | GPT & \xmark & \xmark  & \xmark & Reply | Para.             \\ 
                         & AdapterCL \cite{madotto2020continual} & GPT2 & \xmark & \xmark & Para. & \xmark            \\ 
                         & DEMIX \cite{gururangan2021demix}       & GPT2 | GPT3 & Para. & \xmark  & \xmark & \xmark           \\ 
                         & CLASSIC \cite{ke2021classic}      & BERT       & \xmark & \xmark & Para. & \xmark      \\ 
                         & LLM-CL \cite{ding2024boosting}   & LLaMA   &  \xmark & \xmark & \xmark & Replay | Reg. | Para.      \\ 
                         & CPT \cite{ke2022continualb}      & RoBERTa       & \xmark & \xmark & Para. & \xmark      \\ 
                         & C-PT \cite{zhu2022continual}      & T5       & \xmark & \xmark & Para. & \xmark      \\ 
                         & CL-KD \cite{castellucci-etal-2021-learning}      & BERT       & Reg. & \xmark & \xmark & \xmark      \\ 
                         & PlugLM \cite{cheng2022language}      & BERT       & Para.  & \xmark & \xmark & \xmark      \\ 
                         & Pretr \cite{cossu2022continual}      & BERT | RoBERTa        & \xmark & Para. & \xmark & \xmark      \\ 
                         & AEWC \cite{lee2017toward}     & LSTM-RNNs       & Reg. & \xmark & \xmark & \xmark      \\ 
                         & Continual DAP-training \cite{ke2023continual}      & RoBERTa       & \xmark & Reg. & \xmark & \xmark      \\ 
                         & COPF \cite{zhang2023copf}      & LLaMA       & Replay | Reg. | Para. & \xmark & \xmark & \xmark      \\ 
                         & LAMOL \cite{sun2019lamol}      & GPT2       & Replay & \xmark & \xmark & \xmark      \\ 
                         & RVAE\_LAMOL \cite{wang2022rvae}      & GPT2       & Replay & \xmark & \xmark & \xmark      \\ 
                         & Adapt-Retrieve-Revise \cite{zhang2023reformulating}      & Baichuan       & \xmark & \xmark & \xmark &  RAG      \\ 
                         & Lifelong-MoE \cite{chen2023lifelong}      & GLaM       & \xmark & \xmark & Para. & \xmark      \\ 
                         & CPPO \cite{zhangcppo}      & GPT2       & Reg. & \xmark & \xmark & \xmark      \\ 
                         & EcomGPT-CT \cite{ma2023ecomgpt}       & BLOOM       & \xmark & Replay & \xmark & \xmark      \\ 
                         & S-Prompt \cite{wang2022s}      & CLIP       & \xmark  & \xmark & Para. & \xmark      \\ 
                         & VQACL \cite{zhang2023vqacl} & VL-T5 & Replay | Reg. & \xmark & \xmark & \xmark \\ \hline
\multirow{27}{*}{TIL}    & PP \cite{razdaibiedina2023progressive} & BERT | T5       & \xmark & \xmark & Para. & \xmark            \\ 
                         & CTR \cite{ke2021achieving}      & BERT       & Para. & \xmark & \xmark           & \xmark \\ 
                         & MeLL \cite{wang2021mell} & BERT | RoBERTa & Replay & \xmark & \xmark & \xmark        \\ 
                         & ERDA \cite{qin2022continual}  & Bi-LSTM | BERT       & Replay | Reg. & \xmark & \xmark & \xmark            \\ 
                         & PCLL \cite{zhao2022prompt}  & GPT2       & \xmark  & \xmark  & \xmark & Replay            \\ 
                         
                         & BiHNet-Reg \cite{jin2021learn} & BART | BART-Adapter    & \xmark  & Para.  & \xmark & \xmark            \\ 
                         & ConTinTin \cite{yin2022contintin} & BART       & \xmark & \xmark & \xmark & Replay | InstructionSpeak     \\ 
                         & HMI \cite{maekawa2023generative}  & BERT, GPT2       & Replay & \xmark  & \xmark        & \xmark            \\ 
                         & ACM \cite{zhang2022continual} &  GPT2 & \xmark & Replay | Para. & \xmark & \xmark          \\ 
                         & DYNAINST \cite{mok-etal-2023-large} & BART   & \xmark  & \xmark  & \xmark & Replay | Reg.      \\ 
                         & Conure \cite{10.1145/3404835.3462884} & TCN   & Para.  & \xmark  & \xmark & \xmark      \\ 
                         & TERACON \cite{kim2023task} & NextitNet   & Para.  & \xmark  & \xmark & \xmark      \\ 
                         & ERNIE 2.0 \cite{sun2020ernie} & ERNIE   & \xmark  & \xmark  & Replay & \xmark      \\ 
                         & RecyclableTuning \cite{qin2023recyclable} & RoBERTa   & \xmark  & Replay | Para.  & \xmark & \xmark      \\ 
                         & Conpet \cite{song2023conpet} & LLaMA   & \xmark  & \xmark  & Replay | Para. & \xmark      \\ 
                         & InstructAlign \cite{cahyawijaya2023instruct} & BLOOMZ   & \xmark  & \xmark  & \xmark & Replay      \\ 
                         & Continual-T0 \cite{scialom2022continual} & T0   & \xmark  & \xmark  & \xmark & Replay      \\ 
                         & DynaMind \cite{du2023static} & LLaMA | GPT | Falcon   & Replay | Para.  & \xmark  & \xmark & \xmark      \\ 
                         & ELM \cite{jang2023exploring} & T5   & \xmark  & \xmark  & Para. & \xmark      \\ 
                         & O-LoRA \cite{wang-etal-2023-orthogonal} & LLaMA | Alpaca   & \xmark  & \xmark  & Para. & \xmark      \\ 
                         & COPF \cite{zhang2023copf}      & LLaMA       & Replay | Reg. | Para. & \xmark & \xmark & \xmark      \\ 
                         & SLM \cite{peng2024scalable} & LLaMA, T5, BERT   & Para.  & \xmark  & \xmark & \xmark      \\ 
                         & CTP \cite{zhu2023ctp} & BERT | ViT   & Para.  & \xmark  & - & \xmark      \\ 
                         & ZSCL \cite{zheng2023preventing} & CLIP   & Replay | Reg.  & \xmark  & \xmark & \xmark      \\ 
                         & MoE-Adapters4CL \cite{yu2024boosting} & CLIP   & Para.  & \xmark  & \xmark & \xmark      \\ 
                         & TRIPLET \cite{qian2023decouple} & ALBEF | FLAVA     & \xmark & \xmark & \xmark & Para.  \\
                         & RecAdam \cite{chen2020recall} & BERT | ALBERT     & Reg. & \xmark & \xmark & \xmark \\ \hline
\multirow{16}{*}{CIL}    & EPI \cite{wang2023rehearsal} & BERT       & Para. & \xmark & \xmark & \xmark            \\ 
                         & IDBR \cite{huang2021continual}      & BERT       & Replay & \xmark & \xmark           & \xmark \\ 
                         & PAGeR \cite{varshney2022prompt} & GPT2 & \xmark & \xmark & \xmark & Replay        \\ 
                         & ENTAILMENT \cite{xia2021incremental}    & RoBERTa       & \xmark      & \xmark  & \xmark & prompt-based         \\ 
                         & ExtendNER \cite{monaikul2021continual}  & BERT       & Reg. & \xmark & \xmark & \xmark            \\ 
                         & PLE \cite{li2022continual}  & RoBERTa       & \xmark  & \xmark  & Replay | Para. & \xmark            \\ 
                         & DE\&E \cite{wojcik2023domain} & Distilbert    & \xmark  & Para.  & \xmark & \xmark            \\ 
                         & SRC \cite{liu2019continual} & SIF       & Reg. & \xmark & \xmark & \xmark     \\ 
                         & MoE-Adapters4CL \cite{yu2024boosting}  & TIL       & RandSel | KMeansSel & \xmark  & \xmark        & \xmark            \\ 
                         & VLM-PL \cite{kim2024vlm} & CLIP | Ferret & Replay & \xmark & \xmark & \xmark          \\ 
                         & Adaptation-CLIP \cite{liu2023class} & CLIP   & \xmark  & \xmark  & Para. & \xmark      \\ 
                         & PROOF \cite{zhou2023learning} & CLIP   & Para.  & \xmark  & \xmark & \xmark      \\ 
                         & LGCL \cite{khan2023introducing} & CLIP | ViT   & \xmark  & \xmark  & \xmark & Para.      \\ 
                         & ZSCL \cite{zheng2023preventing} & CLIP   & Replay | Reg.  & \xmark  & \xmark & \xmark      \\
                         & CLAP \cite{jha2024clap4clip} & CLIP   & Replay | Para.  & \xmark  & \xmark & \xmark      \\ 
                         & GMM \cite{cao2024generative} & EVA-CLIP | BLIP2     & Reg. & \xmark & \xmark & \xmark            \\ \hline
\multirow{9}{*}{Online CL}    
                        & MBPA++ \cite{de2019episodic} & BERT & Replay & \xmark & \xmark & \xmark            \\ 
                        & Meta-MBPA++ \cite{wang2020efficient} & BERT & Replay & \xmark & \xmark & \xmark            \\ 
                        & OML-ER \cite{holla2020meta} & BERT & Replay & \xmark & \xmark & \xmark            \\ 
                        & TPEM \cite{geng2021continual} & GRU & Para. & \xmark & \xmark & \xmark            \\ 
                        & MSR \cite{liu2021lifelong1} & BERT & Replay | Reg. & \xmark & \xmark & \xmark            \\ 
                        & ProgModel \cite{shen2019progressive} & RNN & \xmark & Para. & \xmark & \xmark            \\ 
                        & DKR \cite{cui2024continual} & CLIP & Reg. & \xmark & \xmark & \xmark            \\ 
                        & GMM \cite{cao2024generative} &  EVA-CLIP | BLIP2 | MiniGPT-4     & Para. & \xmark & \xmark & \xmark            \\ \hline
\end{tabular}
}
\vspace{-3mm}
\end{table*}

\subsection{Metrics for Continual Pre-training.}
CKL \cite{jang2021towards} introduces FUAR, which quantitatively measures the efficiency of each CKL method. \highlight{It calculates the instances of time-invariant knowledge that a model forgets to learn or update one instance of new knowledge.} 


\begin{table*}[t!]
\centering
\scriptsize
\caption{A summary of the performance of various Foundation LMs. A-RG: Average of ROUGE-1, -2 and -L scores; F1: F1 score; Acc.: Accuracy; MF1: Macro-F1; AMF1: averaged Macro-F1; JGA: Joint Goal Accuracy; SER: Slot Error Rate; AA: average accuracy; AIA: average incremental accuracy; FM: forgetting measure.}
\label{tab:result}
\vspace{-3mm}
\setlength{\tabcolsep}{1.5pt} 
\begin{tabular}{cp{5cm}ccccccccccccc}
\hline
\textbf{Methods}               & \textbf{Dataset} & \textbf{F1} & \textbf{Acc} & \textbf{A-RG} & \textbf{MF1}        & \textbf{JGA}                                                                                                                                                                             & \textbf{SER} & \textbf{BLEU} & \textbf{AA} & \textbf{AIA} & \textbf{FM} & \textbf{FWT} & \textbf{BWT} & \textbf{AMF1}  \\ \hline
LFPT5 \cite{qin2022lfpt}            &       AGNews, Amazon, DBPedia, Yahoo           &       -         &             52.71    &             -     & -   & -                                                            & -           & -    & -           & - & -    & -           & -& -    \\
SLM(T5) \cite{peng2024scalable}            &       AGNews, Amazon, DBpedia, Yahoo           &       -         &             -    &             -     &   -   & -                                                            & -   & -    &     73.10            & -         & -     & -           & -  & - \\
PP(T5) \cite{razdaibiedina2023progressive}            &      AGNews, Amazon, DBpedia, Yahoo           &       -         &             -    &             -     & -   & -                                                            & -   & -    & 75.10           & -         & -     & -           & - & -  \\
MBPA++ \cite{de2019episodic}            &       AGNews, Amazon, DBpedia, Yahoo, Yelp           &       -         &             -    &             -     & -   & -                                                            & -   & -    &  70.60           & -         & -     & -           & - & -  \\ 
IDBR \cite{huang2021continual}            &       AGNews, Amazon, DBpedia, Yahoo, Yelp          &       -         &             -    &             -     &   -   & -                                                            & -   & -    &      73.19            & -         & -     & -           & -  & - \\
EPI \cite{wang2023rehearsal}            &       AGNews, Amazon, DBpedia, Yahoo, Yelp           &       -         &             -    &             -     &   -   & -                                                            & -   & -    &     74.43            & -         & -     & -           & -  & - \\
OML-ER \cite{holla2020meta}            &       AGNews, Amazon, DBpedia, Yahoo, Yelp           &       -         &             -    &             -     & -   & -                                                            & -   & -    &  75.70           & -         & -     & -           & - & -  \\
O-LoRA\cite{wang-etal-2023-orthogonal}            &       AGNews, Amazon, DBpedia, Yahoo, Yelp           &       -         &             -    &             -     &   -   & -                                                            & -   & -    &     75.80            & -         & -     & -           & -  & - \\
LAMOL \cite{sun2019lamol}            &       AGNews, Amazon, DBPedia, Yahoo, Yelp           &       -         &             -    &             -     & -   & -                                                            & -   & -    & 76.50           & -         & -     & -           & -  & - \\
PP(Bert) \cite{razdaibiedina2023progressive}            &       AGNews, Amazon, DBpedia, Yahoo, Yelp           &       -         &             -    &             -     & -   & -                                                            & -   & -    &  77.90           & -         & -     & -           & - & -  \\
Meta-MBPA++ \cite{wang2020efficient}            &       AGNews, Amazon, DBpedia, Yahoo, Yelp           &       -         &             -    &             -     & -   & -                                                            & -   & -    &  77.30           & -         & -     & -           & - & -  \\ 
SLM(Bert) \cite{peng2024scalable}            &       AGNews, Amazon, DBpedia, Yahoo, Yelp           &       -         &             -    &             -     &   -   & -                                                            & -   & -    &     79.10            & -         & -     & -           & -  & - \\
B-CL \cite{ke2021adapting}            &       HL5Domains, Liu3Domains, Ding9Domains, SemEval14           &       -         &             88.29    &             -     & 81.40   & -                                                            & -           & -     & -           & -  & -    & -           & - & -  \\
CLASSIC \cite{ke2021classic}            &       HL5Domains, Liu3Domains, Ding9Domains, SemEval14           &       -         &             90.22    &             -     & 85.12   & -                                                            & -           & -     & -           & -  & -    & -           & - & - \\
CTR \cite{ke2021achieving}            &       HL5Domains, Liu3Domains, Ding9Domains, SemEval14           &       -         &             89.47    &             -     &  83.62   & -                                                            & -   & -    &  -           & -         & -     & -           & -  & - \\
LLM-CT \cite{ding2024boosting}            &       HL5Domains, Liu3Domains, Ding9Domains, SemEval14           &       -         &             94.91    &             -     & 91.43   & -                                                            & -           & -     & -           & -  & -    & -           & - & - \\
PP(Bert) \cite{razdaibiedina2023progressive}            &       MNLI, CB, WiC, COPA, QQP, BoolQ, RTE, IMDB, Yelp, Amazon, SST2, DBpedia, AGNews, Yelp, Amazon, Yahoo           &       -         &             -    &             -     & -   & -                                                            & -   & -    &  69.3           & -         & -     & -           & - & -  \\
O-LoRA\cite{wang-etal-2023-orthogonal}            &       MNLI, CB, WiC, COPA, QQP, BoolQ, RTE, IMDB, Yelp, Amazon, SST2, DBpedia, AGNews, Yelp, Amazon, Yahoo           &       -         &             -    &             -     &   -   & -                                                            & -   & -    &     69.6            & -         & -     & -           & -  & - \\
PP(T5) \cite{razdaibiedina2023progressive}            &       MNLI, CB, WiC, COPA, QQP, BoolQ, RTE, IMDB, Yelp, Amazon, SST2, DBpedia, AGNews, Yelp, Amazon, Yahoo           &       -         &             -    &             -     & -   & -                                                            & -   & -    &  79.5           & -         & -     & -           & -  & - \\
OML-ER \cite{holla2020meta}            &       FewRel           &       -         &             69.5    &             -     & -   & -                                                            & -   & -    &  -           & -         & -     & -           & - & -  \\
DYNAMIC CONPET\cite{song2023conpet}            &       FewRel           &       -         &             88.62    &             -     &   -   & -                                                            & -   & -    &    88.62           & -         & -     & -           & -  & - \\
PAGeR \cite{varshney2022prompt}            &       FewRel           &       -         &             -    &             -     &   -   & -                                                            & -   & -    &      91.30            & -         & -     & -           & -  & - \\
CPPO \cite{zhangcppo}            &       IMDB, HH-RLHF, Reddit TL, DR           &       -         &             -    &             -     & -   & -                                                            & -   & -    & 77.40           & 84.20         & 2.9     & -           & -2.9  & - \\
COPF \cite{zhang2023copf}            &       IMDB, HH-RLHF, Reddit TL, DR           &       -         &             -    &             -     & -   & -                                                            & -   & -    & 86.30           & 85.10         & -0.60     & -           & 0.60 & -  \\
LAMOL \cite{sun2019lamol}            &       SST, QA-SRL, WOZ           &       -         &             -    &             -     & -   & -                                                            & -   & -    & 81.00           & -         & -     & -           & -  & - \\
RVAE\_LAMOL \cite{wang2022rvae}           &       SST, QA-SRL, WOZ           &       -         &             -    &             -     & -   & -                                                            & -   & -    & 81.60           & -         & -     & -           & -  & - \\
PLE(5-shot regime) \cite{li2022continual}            &        CLINC150, ATIS, HWU64, BANKING77, MTOP, SNIPS, LEYZER, MSLU, TOP           &       -         &             -    &             -     &   -   & -                                                            & -   & -    &      85.61            & -         & -     & -           & -  & - \\
PLE(10-shot regime) \cite{li2022continual}            &       CLINC150, ATIS, HWU64, BANKING77, MTOP, SNIPS, LEYZER, MSLU, TOP           &       -         &             -    &             -     &   -   & -                                                            & -   & -    &      88.16            & -         & -     & -           & -  & - \\
MBPA++ \cite{de2019episodic}            &       QuAC, TrWeb, TrWik, SQuAD           &       -         &             -    &             -     & -   & -                                                            & -   & -    &  -           & -         & -     & -           & - & 62.40  \\
Meta-MBPA++ \cite{wang2020efficient}            &       QuAC, TrWeb, TrWik, SQuAD           &       -         &             -    &             -     & -   & -                                                            & -   & -    &  -           & -         & -     & -           & - & 64.90  \\ 
CPT \cite{ke2022continualb}            &       Yelp Restaurant, AI Papers, ACL Papers, AGNews           &       -         &             52.59    &             -     & 46.41   & -                                                            & -           & -     & -           & - & -    & -           & -  & - \\
Continual DAP-training \cite{ke2023continual}            &       Yelp Restaurant, Amazon Phone, Amazon Camera, ACL Papers, AI Papers, PubMed Papers           &       -         &             81.91    &             -     & 77.93   & -                                                            & -   & -    & -           & -         & -     & -           & - & -  \\
DYNAMIC CONPET\cite{song2023conpet}            &       OntoNotes           &       -         &             84.47    &             -     &   -   & -                                                            & -   & -    &  85.83           & -         & -     & -           & -  & - \\
LFPT5 \cite{qin2022lfpt}            &       CoNLL03, OntoNotes           &       47.59         &             -    &             -     & -   & -  & -           & -  & -           & - & -  & -           & -   & -   \\
AdapterCL \cite{madotto2020continual}            &       TM19, TM20, SGD, MultiWoZ           &       -         &             90.50    &             -     & -   & 35.10                                                            & 31.78           & 16.76      & -           & - & -    & -           & - & - \\
RMR\_ DSE \cite{li-etal-2022-overcoming}            &       MultiWoZ-2.0           &       -         &             -    &             -     & -   & -                                                            & 48.79           & 39.86      & -    & -    & -           & -        & - & - \\
DEMIX \cite{gururangan2021demix}            &       In-Car Assistant, Multi-WOZ 2.1, CamRest           &       45.84         &             -    &             -     & -   & -                                                            & -           & 11.96      & -           & - & -    & -           & - & - \\
LFPT5 \cite{qin2022lfpt}            &       CNNDM, WikiHow, XSum           &       -         &             -    &             17.05     & -   & -                                                            & -       & -           & -      & -  & -    & -           & -   & -  \\
C-PT \cite{zhu2022continual}            &       Schema-Guided Dialog dataset (SGD)           &       -         &             -    &             -     & -   & 61.20                                                            & -           & -   & -    & -           & -   & 13.70           & 0.50  & - \\
Lifelong-MoE \cite{chen2023lifelong}            &       TriviaQA           &       20.22         &             -    &             -     & -   & -                                                            & -   & -    & -           & -         & -     & -           & -  & - \\
Lifelong-MoE \cite{chen2023lifelong}            &       WMT16           &       -         &             -    &             -     & -   & -                                                            & -   &  19.16    & -           & -         & -     & -           & - & -  \\
CTR \cite{ke2021achieving}            &       10 DSC datasets           &       -         &             89.31    &             -     &   88.75   & -                                                            & -   & -    &  -           & -         & -     & -           & -  & - \\
CTR \cite{ke2021achieving}            &       20News           &       -         &             95.25    &             -     &   95.23   & -                                                            & -   & -    &  -           & -         & -     & -           & - & -  \\
MeLL \cite{wang2021mell}            &       TaskDialog-EUIC           &       -         &             -    &             -     &   -   & -                                                            & -   & -    &  93.79           & -         & -     & -           & -  & 93.42 \\
MeLL \cite{wang2021mell}            &       Hotline-EUIC           &       -         &             -    &             -     &   -   & -                                                            & -   & -    &  96.73           & -         & -     & -           & -  & 93.41 \\
PCLL \cite{zhao2022prompt}            &       HWU, BANKING, CLINC, SNIPS, TOP           &       -         &             -    &             -     &   -   & -                                                            & -   & -    &  90.25           & -         & -     & -           & -  & - \\
PCLL \cite{zhao2022prompt}            &       SNIPS, AITS, DSTC, MIT-MOVIE, MIT-RESTAURANT           &       -         &             -    &             -     &   -   & -                                                            & -   & -    &  -           & -         & -     & -           & -  & 74.48 \\
SLM(LLaMA) \cite{peng2024scalable}            &       Medical, MMLU, Finance           &       -         &             -    &             -     &   -   & -                                                            & -   & -    &     82.3            & -         & -     & -           & -  & - \\
EPI \cite{wang2023rehearsal}            &       Web of Science (WOS)           &       -         &             -    &             -     &   -   & -                                                            & -   & -    &     77.83            & -         & -     & -           & -  & - \\
IDBR \cite{huang2021continual}            &       Yahoo, AGNews, Yelp           &       -         &             -    &             -     &   -   & -                                                            & -   & -    &      72.53            & -         & -     & -           & -  & - \\
\hline
\end{tabular}
\vspace{-3mm}
\end{table*}

\subsection{Online CL-Specific Metrics.}
Near-future accuracy (NFA) \cite{al2023rapid} is introduced as a novel evaluation metric for OCL problems. Unlike traditional evaluation methods that assess models on immediately subsequent samples, NFA evaluates models on samples slightly further into the future, using a minimal shift \( S \). Such operation can mitigate label correlation effects, which can adversely impact the accuracy of model \highlight{adaptability assessments. 
Yogatama et al.} \cite{yogatama2019learning} proposed a novel online codelength, inspired by prequential encoding \cite{blier2018description}, to quantify how quickly an existing model can adapt to a new task. 

\section{Analysis and Discussion}
\highlight{In this section, we compare the performance and characteristics of typical foundation model-based continual learning methods (Table \ref{table: comparison of CL} and Table \ref{tab:result}).
}

\highlight{
The effectiveness of each continual learning method is shaped by the specific task environment, resource constraints, and application goals. Replay-based methods like LAMOL, Meta-MBPA++, PAGeR, and COPF excel in scenarios where memory efficiency and maintaining a balance between old and new knowledge are essential. These methods strike a delicate balance between retaining previously acquired knowledge and adapting to new tasks. Parameter isolation is key to minimizing interference between highly distinct tasks. By allocating dedicated parameters to each task, these approaches prevent overlap and ensure that task-specific learning is preserved. Techniques such as AdapterCL, B-CL, and EPI are particularly effective in environments with highly variable tasks, where reducing interference is critical. Regularization techniques, on the other hand, enhance stability in structured task settings. These methods constrain model updates to protect knowledge from previous tasks, making them ideal in situations with well-defined task boundaries. Regularization-based approaches like RMR\_DSE, CLASSIC, and EWC mitigate catastrophic forgetting by limiting model changes, thereby preserving prior learning. These methods are particularly effective in domain-incremental learning, where tasks evolve gradually, and long-term stability is vital.
}

\highlight{
First, most continual learning (CL) methods are built on PLMs like BERT, RoBERTa, and BART, or smaller-scale LLMs such as LLaMA. Models leveraging LLMs often outperform PLM-based ones, thanks to their greater capacity for handling complex tasks and enhanced adaptability. Second, the majority of CL methods are tailored for offline settings, where the full target dataset is assumed to be available for training. However, in real-world scenarios, data typically arrives in a streaming fashion, making it challenging to anticipate distribution shifts and effectively manage dynamic data. Third, a significant hurdle in the field is the lack of unified benchmarks and standardized metrics for evaluating CL algorithms. This fragmentation complicates the comparison of methods across tasks and domains. To propel the field forward, there is a critical need for benchmark datasets and robust metrics that can consistently gauge the performance of CL models, particularly in dynamic, data-streaming environments. These benchmarks should span diverse tasks and modalities to ensure comprehensive evaluation across various domains.
}

\section{Challenges and Further Work}
\label{sect: challenges and further work}
\paragraph{Autonomous Continual Learning.} 
Most existing studies in the domain of continual learning assume static datasets with known distributions in a relatively closed environment. 
Moreover, these studies mainly focus on simple tasks (e.g., text classification, sentiment analysis and intent classification) with clear labels. 
These assumptions do not hold in real-world applications, where environments continually evolve and introduce novel stimuli.
\highlight{A key challenge is developing continual learning models that can autonomously detect and adapt to data distribution shifts, thereby improving the applicability of AI in dynamic real-world scenarios.}
Liu et al. \cite{liu2023ai} recently proposed the SOLA framework to address these limitations by facilitating autonomous adaptation in AI systems. Despite this progress, significant challenges remain in enabling these systems to adjust to new, dynamic environments without ongoing human oversight. 

\vspace{-2mm}
\paragraph{Bridging Machine Learning and Cognitive Science in Continual Learning.} \highlight{While continual learning aims to mimic human-like learning, a gap persists in how well these methods align with cognitive science findings on human continuous learning. In human cognition, processes such as rehearsal, memory consolidation, and adaptive forgetting are key to learning in dynamic environments. In contrast, machine learning models often face challenges like catastrophic forgetting, a phenomenon without a clear biological equivalent. Bridging these fields could enhance the design of continual learning algorithms by drawing on cognitive science insights, potentially leading to more robust, adaptive, and human-like learning systems.
}

\vspace{-2mm}
\paragraph{Learning Knowledge from Conversation.}
Traditional AI systems are typically trained on static data sets, which starkly contrasts with human conversational learning that dynamically updates knowledge through interaction \cite{liu2021lifelong}. The challenge for AI lies in transitioning from static data learning to more dynamic, conversational engagements. 
The future direction in this area could involve the development of models that mimic human conversational learning processes, capable of context adaptation, new concept inference, and dynamic knowledge application within ongoing interactions.

\vspace{-2mm}
\paragraph{Multi-modal Continual Learning.}
Continual learning research has predominantly concentrated on natural language processing tasks such as sentiment analysis and text classification. Recent studies have begun exploring basic multi-modal tasks, such as text-to-image retrieval, text-image classification, and visual question answering. 
The integration of diverse data types—textual, visual, and auditory—poses a substantial challenge. 
Future studies should expand to more complex multi-modal datasets and strive to devise methodologies that effectively synthesize these varied modalities, thereby enhancing the model's capability to maintain continuous learning across different sensory inputs.

\vspace{-2mm}
\paragraph{Privacy Protection in Continual Learning.}
Privacy protection in continual learning systems poses a significant challenge, particularly as these systems are designed to continuously update and refine their models based on incoming data streams. Unlike traditional static machine learning models, continual learning systems frequently access and process sensitive data across different contexts and time periods, raising substantial concerns about data confidentiality and user privacy. Effective privacy-preserving mechanisms must be integrated into the architecture of these systems to ensure that they do not inadvertently expose or misuse personal data. Techniques such as differential privacy \cite{dwork2006differential}, federated learning \cite{zhang2021survey}, and secure multi-party computation \cite{goldreich1998secure} offer promising solutions by allowing models to learn from decentralized data sources without needing to access the actual data directly. 

\vspace{-2mm}
\paragraph{Continual Alignment.} \highlight{In dynamic real-world environments, human preferences and requirements are ever-evolving, making continual alignment crucial for maintaining AI system effectiveness. Traditional alignment methods often assume static preferences, which is insufficient in changing contexts. Integrating Reinforcement Learning from Human Feedback (RLHF) into continual learning presents a unique challenge, as models must adapt to new preferences while retaining previously learned ones. Future research should prioritize developing methods like CPPO and COPF that balance integrating new feedback with preserving past alignment, ensuring long-term adaptability to human needs.}

\vspace{-2mm}
\paragraph{Robust Continual Learning.}
The existing studies mainly focus on designing a continual learning model to improve the performance of forgetting and transferring with various metrics while the robustness of continual learning systems is not well studied. It is critical, especially in applications where safety and reliability are paramount. The main challenges include evaluating the robustness of these systems against adversarial attacks or when faced with drastically changing environments. 
Future research could focus on developing evaluation metrics for robustness in continual learning and designing systems that maintain performance reliability over time despite environmental changes.

\vspace{-2mm}
\paragraph{Large-Scale and High-Quality Datasets and Benchmarks.}  
As discussed in Section \ref{sect: Datasets}, most of the datasets are constructed by merging the existing datasets. This often results in datasets that lack diversity and real-world complexity, which hampers the development of robust and adaptable continual learning models. 
The creation of large-scale, high-quality datasets that accurately reflect real-world complexities represents a critical challenge. Moving forward, the development of such datasets and benchmarks will be essential not only for assessing the efficacy of continual learning algorithms but also for pushing the limits of what these algorithms can achieve in practical settings.

\section{Conclusions}
\label{sect: Conclusions}
This survey provides an in-depth exploration of continual learning (CL) methodologies tailored for foundation language models (LMs), such as pre-trained language models (PLMs), large language models (LLMs), and vision-language models (VLMs). 
By integrating the dynamic adaptability of CL with the robust foundational capabilities of LMs, this field promises to significantly advance the state of artificial intelligence. 
We categorize existing research into offline and online continual learning paradigms, offering a clear distinction between the settings and methodologies used within these frameworks. Offline CL is discussed in terms of domain-incremental, task-incremental, and class-incremental learning. Meanwhile, online CL is analyzed with a focus on the delineation between hard and blurry task boundaries, providing insights into how these approaches handle real-time data streams. 
Our review of the literature not only clarifies the current landscape of CL approaches for foundation LMs but also emphasizes the innovative integration of continual pre-training, parameter-efficient tuning, and instruction tuning methods that are specifically designed to leverage the vast capabilities of foundation LMs. Furthermore, we highlight the main characteristics of datasets used in this domain and the metrics that effectively measure both the mitigation of catastrophic forgetting and the enhancement of knowledge transfer.
This work hopes to inspire further research that will ultimately lead to more robust, efficient, and intelligent systems capable of lifelong learning.
\bibliographystyle{ACM-Reference-Format}
\bibliography{sample-base}

\newpage
\clearpage
\appendix

\section{Details of Metrics}

\subsection{Overall Performance.} 
Moreover, Chaudhry et al. \cite{chaudhry2018efficient} devise a metric known as Learning Curve Area (LCA), which quantifies the speed of learning in a model. It first defines an average \( b\)-shot performance, where \( b \) represents the number of mini-batches, subsequent to the completion of training across all \( T \) tasks as follows:

\begin{equation}
Z_b = \frac{1}{N} \sum_{i=1}^{N} R_{N,i}
\end{equation}

\( LCA \) at \( \beta \) is the area of the convergence curve \( Z_b \) as a function of \( b \in [0, \beta] \):

\begin{equation}
LCA_{\beta} = \frac{1}{\beta + 1} \int_{0}^{\beta} Z_b \, db = \frac{1}{\beta + 1} \sum_{b=0}^{\beta} Z_b
\end{equation}

The Learning Curve Area (LCA) provides insights into model learning dynamics. \(LCA_{0}\) measures the average \( 0 \)-shot performance, similar to forward transfer (\cite{lopez2017gradient}). \(LCA_{\beta}\), quantifying the area under the \(Z_b\) curve, evaluates both average \( 0 \)-shot performance and learning speed. Although two models may achieve similar \(Z_b\) or \(A_T\) values, they can differ significantly in \(LCA_{\beta}\) due to variations in learning rates. This metric is crucial for identifying models that quickly learn from few examples, particularly when \( \beta \) is small.

Qin et al. \cite{qin2022elle} propose two metrics designed to evaluate pre-trained language models (PLMs) based on their performance within learned domains: Average Perplexity ($AP$) and Average Increased Perplexity ($AP^{+}$). The aforementioned metrics are utilized to assess key capabilities of PLMs, such as instruction following and safety, as discussed in Wang et al. \cite{wang2023trace}.

\subsection{Memory Stability.} 
Chaudhry et al. \cite{chaudhry2018riemannian} introduce the Forgetting Measure (FM), a metric designed to quantify the extent of forgetting a model experiences for a specific task. The forgetting for a given task \( T_j \) after sequential training on tasks up to \( T_N \) is quantified as the difference between the highest proficiency (\( max(R_{l,j}) \)) achieved on task \( T_j \) during initial training and its proficiency (\( R_{N,j} \)) after subsequent learning phases:

\begin{equation}
f_{j} = \max_{l \in \{1, \ldots, N-1\}} (R_{l,j} - R_{N,j}), \quad \forall j < N.
\end{equation}
For the purpose of quantifying forgetting in previous tasks, the function \( f_{j} \) is defined within the interval \([-1, 1]\) for \( j < N \).

Furthermore, to account for the number of tasks previously encountered, the Forgetting Measure (FM) at the \(N\)-th task represents the mean level of forgetting across all preceding tasks:

\begin{equation}
FM = \frac{1}{N-1} \sum_{j=1}^{N-1} f_{j}
\end{equation}
A lower FM indicates better retention of previous tasks. Here, the \textit{expansion} or \( R_{j,j} \) serves as a more effective quantifier of retained knowledge concerning past tasks, as opposed to using \textit{max}. Nonetheless, \textit{max} remains a valuable estimator for assessing the extent of forgetting that occurs throughout the learning process.

Davari et al. \cite{davari2022probing} propose a method named linear probes (LP) to assess representation forgetting. This approach measures the effectiveness of learned representations via an optimal linear classifier trained on the frozen activations of a base network. Representation forgetting is quantified by evaluating the change in Language Processing (LP) performance before and after the introduction of a new task. Formally, for each model (\( f_{\theta_i} \)) at time step \( i \) of a task sequence, the classifier (\( W^*_i \)) is optimized as: \( W^*_i = \arg\min_{W_i} \mathcal{L}(W_i; f_{\theta_i}(X_i), Y_i) \), where \( \mathcal{L} \), \( X_i \), and \( Y_i \) represent the objective function, input data, and labels for task \( i \), respectively. The degree of representational forgetting between two model states, \( \theta_a \) and \( \theta_b \), where \( \theta_b \) is derived later in the sequence, is evaluated by calculating the difference in scores: \( Score(W_a f_{\theta_a}(X_a), Y_a) - Score(W_b f_{\theta_b}(X_a), Y_a) \), where \( Score \) represents the performance metric, such as accuracy, on the task.

Kemker et al. \cite{kemker2018measuring} introduce three metrics designed to CF: \( \Omega_{\text{base}} \), \( \Omega_{\text{new}} \), and \( \Omega_{\text{all}} \). \( \Omega_{\text{base}} \) assesses retention of initial learning, \( \Omega_{\text{new}} \) measures recall of new tasks, and \( \Omega_{\text{all}} \) evaluates overall proficiency in maintaining old knowledge and acquiring new information.

\begin{equation}
\Omega_{\text{base}} = \frac{1}{N-1} \sum_{i=2}^{N} \frac{\alpha_{\text{base},i}}{\alpha_{\text{ideal}}}
\end{equation}
\begin{equation}
\Omega_{\text{new}} = \frac{1}{N-1} \sum_{i=2}^{N} \frac{\alpha_{\text{new},i}}{\alpha_{\text{ideal}}}
\end{equation}
\begin{equation}
\Omega_{\text{all}} = \frac{1}{N-1} \sum_{i=2}^{N} \frac{\alpha_{\text{all},i}}{\alpha_{\text{ideal}}}
\end{equation}
where \( N \) represents the total number of sessions, \( \alpha_{\text{new},i} \) is the test accuracy after learning session \( i \), \( \alpha_{\text{base},i} \) denotes the accuracy on the initial session after \( i \) sessions, and \( \alpha_{\text{all},i} \) refers to the test accuracy across all test data for classes encountered up to point \( i \). The ideal performance (\( \alpha_{\text{ideal}} \)) is defined as the offline MLP accuracy on the base set. To facilitate comparative analysis across different datasets, \( \Omega_{\text{base}} \) and \( \Omega_{\text{all}} \) are normalized by \( \alpha_{\text{ideal}} \). Consequently, unless a model surpasses \( \alpha_{\text{ideal}} \), normalized results will range from 0 to 1, enabling consistent cross-dataset comparisons.

Additionally, researchers \cite{koh2022online} devise a novel metric, termed the Knowledge Loss Ratio (KLR), quantifies knowledge degradation using principles from information theory.

\subsection{Learning Plasticity.}
Intransigence measure (IM), as defined by Chaudhry et al. \cite{chaudhry2018riemannian}, quantifies a model's inability to learn new tasks. This measure is calculated by comparing the performance difference of a task when trained jointly with other tasks versus when trained in a continual learning setting. Then  the intransigence for the \( N \)-th task can be defined as:

\begin{equation}
IM = R^*_N - R_{N,N},
\end{equation}
where \( R^*_N \) represents the accuracy achieved on the held-out dataset of the \( N \)-th task, \( R_{N,N} \) indicates the accuracy on the \( N \)-th task upon completion of training in an incremental sequence up to and including task \( N \). Note, \( IM_N \in [-1, 1] \), and lower values indicate superior performance.

\subsection{Metrics for Continual Pre-training.}
CKL \cite{jang2021towards} introduces a novel metric, named FUAR (FORGOTTEN / (UPDATED + ACQUIRED) RATIO), which quantitatively measures the efficiency of each CKL method. It calculates the number of instances of time-invariant knowledge that a model forgets in order to learn or update one instance of new knowledge. When FUAR is equal to 1.0, it signifies an equilibrium where one time-invariant knowledge instance is forgotten on average to obtain a new or updated knowledge instance. Formally, FUAR is defined as:
\begin{equation}
Eq_1 = \sum_{i=0}^{N-1} \max(0, \text{Gap}(T_i^F, D_i, D_N)) \mathbbm{1}_{\{T_i^F \neq n.d.\}}
\end{equation}
\begin{equation}
\begin{split}
        Eq_2=\sum_{i=0}^{N-1}  \max(0, \text{Gap}(T_B^U, D_N, D_i)) \mathbbm{1}_{\{T_i^F \neq n.d.\}} 
        \\+ \max(0, \text{Gap}(T_N^A, D_N, D_i)) \mathbbm{1}_{\{T_i^F \neq n.d.\}} 
\end{split}
\end{equation}

\begin{equation}
    FUAR(\mathbbm{T}^F,T_N^U,T_N^A) = \left\{
    \begin{aligned}
      &\frac{Eq_1}{Eq_2} \quad &&\text{if } denominator > 0 \\
      &no\;gain \quad &&\text{otherwise }
    \end{aligned}
  \right.
\end{equation}
where \( T \) represents an arbitrary task, and \( (D_i)_{i=0}^N \) is a sequence of corpora for LM pretraining. Gap(\( T, D_a, D_b \)) is \( \text{Score}(T) \) of \( LM_a \) - \( \text{Score}(T) \) of \( LM_b \), where \( LM_a \) is pretrained on \( D_a \). \( \mathbbm{T}^F = (T_i^F)_{i=0}^{N-1} \) measures forgetting of invariant-knowledge from \( (D_i)_{i=0}^{N-1} \). If no task is from \( D_i \), \( T_i^F \) is "n.d." (not defined). \( T_N^U \) and \( T_N^A \) from \( D_N \) measure update and acquisition of new knowledge, respectively.

\subsection{Online CL-Specific Metrics.}

Near-future accuracy (NFA) \cite{al2023rapid} is introduced as a novel evaluation metric for OCL problem. Unlike traditional evaluation methods that assess models on immediately subsequent samples, NFA evaluates models on samples slightly further into the future, using a minimal shift \( S \). Such operation can mitigate label correlation effects, which can adversely impact the accuracy of model adaptability assessments. The smallest shift \( S \) is selected to ensure that the test sample aligns closely with the distribution of recently observed training data. The calculation of NFA involves first checking if the model correctly predicts the label of a future sample, which can be expressed as $a_t =  \mathbbm{1}  \{f_{\theta_t}(x_{t+1+S}) = y_{t+1+S}\}$. Subsequently, the running average is updated using the formula \( A^{RA}t = \frac{1}{t} (A^{RA}{t-1} \cdot (t - 1) + a_t) \).

Yogatama et al. \cite{yogatama2019learning} proposed a novel online codelength ($\ell(D)$), inspired by prequential encoding \cite{blier2018description}, to quantify how quickly an existing model can adapt to a new task. 
\begin{equation}
    \ell(D) = \log_2 |y| - \sum_{i=2}^{N} \log_2 p(y_i | x_i; \theta_{D_{i-1}})
\end{equation}
where \( |Y| \) is the number of possible labels (classes), and \( \theta_{D_i} \) represents a particular subset of the dataset \( D \). Similar to the approach in Latent Contextual Allocation (LCA) \cite{chaudhry2018efficient}, the concept of \emph{online codelength} is associated with the area under the learning curve.

\end{document}